\documentclass{article}

\usepackage{microtype}
\usepackage{graphicx}
\usepackage{subfigure}
\usepackage{booktabs} 
\usepackage{mathrsfs}
\usepackage{amsmath,amsfonts}
\usepackage{enumerate}  

\newtheorem{theorem}{\textbf{Theorem}}[section]
\newtheorem{definition}{\textbf{Definition}}[section]
\newtheorem{lemma}{\textbf{Lemma}}[section]

\usepackage{hyperref}

\usepackage[accepted]{icml2018}
\ifodd 0
\newcommand{\rev}[1]{{\color{blue}#1}} 
 
\newcommand{\com}[1]{\textbf{\color{red}(COMMENT: #1)}} 
\newcommand{\mcom}[1]{\textbf{\color{purple}(MahdiCom: #1)}} 
\newcommand{\clar}[1]{\textbf{\color{green}(NEED CLARIFICATION: #1)}}

\else
\newcommand{\rev}[1]{#1}

\newcommand{\com}[1]{}
\newcommand{\mcom}[1]{}

\newcommand{\clar}[1]{}

\fi

\icmltitlerunning{Improving the Privacy and Accuracy of ADMM-Based Distributed Algorithms}

\begin{document}

\twocolumn[
\icmltitle{Improving the Privacy and Accuracy of ADMM-Based Distributed Algorithms }



\icmlsetsymbol{equal}{*}

\begin{icmlauthorlist}
\icmlauthor{Xueru Zhang}{ed}
\icmlauthor{Mohammad Mahdi Khalili}{ed}
\icmlauthor{Mingyan Liu }{ed}
\end{icmlauthorlist}

\icmlaffiliation{ed}{Department of Electrical Engineering and Computer Science, University of Michigan, Ann Arbor, Michigan, USA}

\icmlcorrespondingauthor{Xueru Zhang}{xueru@umich.edu}
\icmlcorrespondingauthor{Mohammad Mahdi Khalili}{khalili@umich.edu}
\icmlcorrespondingauthor{Mingyan Liu}{mingyan@umich.edu}
\icmlkeywords{Distributed Machine Learning, Security and Privacy, Optimization, ADMM}

\vskip 0.3in
]

\printAffiliationsAndNotice{ }  

\begin{abstract}

Alternating direction method of multiplier (ADMM) is a popular method used to design distributed versions of a machine learning algorithm, whereby local computations are performed on local data with the output exchanged among neighbors in an iterative fashion. During this iterative process the leakage of data privacy arises. A differentially private ADMM was proposed in prior work \cite{zhang2017} where only the privacy loss of a single node during one iteration was bounded, a method that makes it difficult to balance the tradeoff between the utility attained through distributed computation and privacy guarantees when considering the total privacy loss of all nodes over the entire iterative process.  We propose a perturbation method for ADMM where the perturbed term is correlated with the penalty parameters; this is shown to improve the utility and privacy simultaneously.  The method is based on a modified ADMM where each node  independently determines its own penalty parameter in every iteration and decouples it from the dual updating step size.  The condition for convergence of the modified ADMM and the lower bound on the convergence rate are also derived.

%
%

\end{abstract}
\section{Introduction}\label{Section1}
Distributed machine learning is crucial for many settings where the data is possessed by multiple parties or when the quantity of data prohibits processing at a central location. It helps to reduce the computational complexity, improve both the robustness and the scalability of data processing. 
In a distributed setting, multiple entities/nodes collaboratively work toward a common optimization objective through an interactive process of local computation and message passing, which ideally should result in all nodes converging to a global optimum. 
%
Existing approaches to decentralizing an optimization problem primarily consist of subgradient-based algorithms \cite{nedic2008,nedic2009,lobel2011}, ADMM-based algorithms \cite{wei2012,ling2014,shi2014,zhang2014,ling2016}, and composite of subgradient and ADMM \cite{bianchi2014}. It has been shown that ADMM-based algorithms can converge at the rate of $O(\frac{1}{k})$ while subgradient-based algorithms typically converge at the rate of $O(\frac{1}{\sqrt{k}})$, where $k$ is the number of iterations \cite{wei2012}.  
In this study, we will solely focus on ADMM-based algorithms.  

The information exchanged over the iterative process gives rise to privacy concerns if the local training data is proprietary to each node, especially when it contains sensitive information such as medical or financial records, web search history, and so on. \rev{It is therefore highly desirable to ensure such iterative processes are privacy-preserving.} 

A widely used notion of privacy is the $\varepsilon$-differential privacy; 
it is generally achieved by perturbing the algorithm such that the probability distribution of its output is relatively insensitive to any change to a single record in the input \cite{Dwork2006}.  
Several differentially private distributed algorithms have been proposed, including \cite{hale2015,huang2015,han2017,zhang2017,bellet2017}. While a number of such studies have been done for (sub)gradient-based algorithms, the same is much harder for ADMM-based algorithms due to its computational complexity stemming from the fact that each node is required to solve an optimization problem in each iteration. To the best of our knowledge, only \cite{zhang2017} applies differential privacy to ADMM, where the noise is either added to the dual variable (\textit{dual variable perturbation}) or the primal variable (\textit{primal variable perturbation}) in ADMM updates. However, \cite{zhang2017} could only bound the privacy loss of a single iteration. Since an attacker can potentially use all intermediate results to perform inference, the privacy loss accumulates over time through the iterative process. It turns out that the tradeoff between the utility of the algorithm and its privacy preservation over the entire computational process becomes hard using the existing method.  

In this study we propose a perturbation method that could simultaneously improve the \rev{accuracy} and privacy for ADMM. We start with a modified version of ADMM whereby each node independently decides its own penalty parameter in each iteration; it may also differ from the dual updating step size. For this modified ADMM we establish conditions for convergence and quantify the lower bound of the convergence rate.  We then present a penalty perturbation method to provide differential privacy. 
Our numerical results show that under this method, by increasing the penalty parameter over iterations, we can achieve stronger privacy guarantee as well as better algorithmic performance, i.e., more stable convergence and higher accuracy.

The remainder of the paper is organized as follows. We present problem formulation and definition of differential privacy and ADMM in Section \ref{sec:prelim} and a modified ADMM algorithm along with its convergence analysis in Section \ref{sec:m-admm}.  A private version of this ADMM algorithm is then introduced in Section \ref{sec:private} and numerical results in Section \ref{sec:numerical}.  Discussions are given in Section \ref{sec:discuss} and Section \ref{sec:conclude} concludes the paper.

\section{Preliminaries }\label{sec:prelim}
\subsection{Problem Formulation}
Consider a connected network\footnote{A connected network is one in which every node is reachable (via a path) from every other node.} given by an undirected graph $G(\mathscr{N},\mathscr{E})$, which consists of a set of nodes $\mathscr{N} = \{1,2,\cdots,N\}$ and a set of edges $\mathscr{E} = \{1,2,\cdots,E\}$. Two nodes can exchange information if and only if they are connected by an edge. Let $\mathscr{V}_i$ denote node $i$'s set of neighbors, excluding itself. A node $i$ contains a dataset $D_i = \{(x_{i}^n,y_{i}^n) | n = 1,2,\cdots,B_i \}$, where $x_{i}^n \in \mathbb{R}^d$ is the feature vector representing the $n$-th sample belonging to $i$, $y_{i}^n \in \{-1,1\}$ the corresponding label, and $B_i$ the size of $D_i$. 

Consider the regularized empirical risk minimization (ERM) problems for binary classification defined as follows:
\begin{equation}\label{eq:prelimi_1}
\min_{f_c}O_{ERM}(f_{c},D_{all}) = \sum_{i=1}^{N}\dfrac{C}{B_i}\sum_{n=1}^{B_i} {\mathscr{L}}(y_{i}^n f_c^Tx_{i}^n ) + \rho R(f_c) 
\end{equation}
where $C \leq B_i$ and $\rho>0$ are constant parameters of the algorithm, the loss function $\mathscr{L}(\cdot)$ measures the accuracy of classifier, and the regularizer $R(\cdot)$ helps to prevent overfitting. The goal is to train a (centralized) classifier $f_c\in \mathbb{R}^d$ over the union of all local datasets $D_{all} = \cup_{i \in \mathscr{N}} D_i$ in a distributed manner using ADMM, while providing privacy guarantee for each data sample \footnote{\rev{The proposed penalty perturbation method is not limited to classification problems. It can be applied to general ADMM-based distributed algorithms since the convergence and privacy analysis in Section \ref{sec:m-admm} \& \ref{sec:private} remain valid.}}.
\subsection{Conventional ADMM}
To decentralize \eqref{eq:prelimi_1}, let $f_i$ be the local classifier of each node $i$. To achieve consensus, i.e., $f_1 = f_2 = \cdots = f_N$, a set of auxiliary variables $\{w_{ij} | i \in \mathscr{N}, j \in \mathscr{V}_i\}$ are introduced for every pair of connected nodes.  As a result, \eqref{eq:prelimi_1} is reformulated equivalently as: 
 \begin{equation}\label{eq:prelimi_2}
\begin{aligned}
& \min_{\rev{\{f_i\},\{w_{ij}\}}} 
& &\tilde{O}_{ERM}(\{f_i\}_{i=1}^N,D_{all})  = \sum_{i=1}^{N}O(f_i,D_i) \\
&\text{   s.t.} 
& & f_i = w_{ij}, w_{ij} = f_j, \ \ \ i \in \mathscr{N}, j \in \mathscr{V}_i
\end{aligned}
\end{equation}
\rev{where  $O(f_i,D_i) = \dfrac{C}{B_i}\sum_{n=1}^{B_i} {\mathscr{L}}(y_{i}^n f_i^Tx_{i}^n ) + \dfrac{\rho}{N} R(f_i)$. 
The objective in \eqref{eq:prelimi_2} can be solved using ADMM.  
Let $\{f_i\}$ be the shorthand for $\{f_i\}_{i\in \mathscr{N}}$; let
$\{w_{ij},\lambda_{ij}^k\}$ be the shorthand for $\{w_{ij},\lambda_{ij}^k\}_{i \in \mathscr{N},j \in \mathscr{V}_i, k \in \{a,b\}}$, where $\lambda_{ij}^a$, $\lambda_{ij}^b$ are dual variables corresponding to equality constraints $f_i = w_{ij}$ and $w_{ij}=f_j$ respectively. 
Then the augmented Lagrangian is as follows: }
\begin{eqnarray}\label{eq:prelimi_3}
L_\eta (\{f_i\},\{w_{ij},\lambda_{ij}^k\}) = \sum_{i=1}^NO(f_i,D_i)\nonumber\\ +  \sum_{i=1}^N\sum_{j \in \mathscr{V}_i}(\lambda_{ij}^a)^T(f_i-w_{ij})+\sum_{i=1}^N\sum_{j \in \mathscr{V}_i}(\lambda_{ij}^b)^T(w_{ij}-f_j)\\
+ \sum_{i=1}^N\sum_{j \in \mathscr{V}_i}\dfrac{\eta}{2} (||f_i-w_{ij}||_2^2+||w_{ij}-f_j||_2^2)~.\nonumber
\end{eqnarray}
In the $(t+1)$-th iteration, the ADMM updates consist of the following:
\begin{eqnarray}
f_i(t+1) = \underset{f_i}{\text{argmin}}\ L_\eta (\{f_i\},\{w_{ij}(t),\lambda_{ij}^k(t)\})~;\label{eq:prelimi_4}\\
w_{ij}(t+1) = \underset{w_{ij}}{\text{argmin}}\ L_\eta (\{f_i(t+1)\},\{w_{ij},\lambda_{ij}^k(t)\})~;\label{eq:prelimi_5}\\
\lambda_{ij}^a(t+1) = \lambda_{ij}^a(t) + \eta(f_i(t+1)-w_{ij}(t+1))~;\label{eq:prelimi_6}\\
\lambda_{ij}^b(t+1) = \lambda_{ij}^b(t) + \eta(w_{ij}(t+1)-f_j(t+1))~.\label{eq:prelimi_7}
\end{eqnarray}
Using Lemma 3 in \cite{forero2010}, {if dual variables $\lambda_{ij}^a(t)$ and $\lambda_{ij}^b(t)$ are initialized to zero for all node pairs $(i,j)$, then $\lambda_{ij}^a(t) = \lambda_{ij}^b(t)$ and $\lambda_{ij}^k(t) = -\lambda_{ji}^k(t)$ will hold for all iterations with $k \in \{a,b\}, i \in \mathscr{N}, j \in \mathscr{V}_i$.}

Let $\lambda_{i}(t) = \sum_{j \in \mathscr{V}_i}\lambda_{ij}^a(t) = \sum_{j \in \mathscr{V}_i}\lambda_{ij}^b(t)$, then the ADMM iterations \eqref{eq:prelimi_4}-\eqref{eq:prelimi_7} can be simplified as:
\begin{eqnarray}
f_i(t+1) = \underset{f_i}{\text{argmin}} \{ O(f_i,D_i) + 2\lambda_i(t)^Tf_i \nonumber \\
+  \eta \sum_{j \in \mathscr{V}_i}||\dfrac{1}{2}(f_i(t)+f_j(t))-f_i||_2^2~ \}~; \label{eq:prelimi_8} \\ 
\lambda_{i}(t+1) = \lambda_{i}(t) +  \dfrac{\eta}{2}\sum_{j \in \mathscr{V}_i}(f_i(t+1)-f_j(t+1))~. \label{eq:prelimi_9}
\end{eqnarray}
\subsection{Differential Privacy}
Differential privacy \cite{Dwork2006} \rev{can be used to} measure the privacy risk of each individual sample in the dataset quantitatively. \rev{Mathematically, a randomized algorithm $ \mathscr{A}(\cdot)$ taking a dataset as input satisfies $\varepsilon$-differential privacy if for any two datasets $D$, $\hat{D}$ differing in at most one data point, and for any set of possible outputs $S \subseteq \text{range}(\mathscr{A})$,  $\text{Pr}(\mathscr{A}(D) \in S)\leq \exp(\varepsilon) \text{Pr}(\mathscr{A}(\hat{D}) \in S)$ holds.}
We call two datasets differing in \rev{at most} one data point as neighboring datasets. 
The above definition suggests that for a sufficiently small $\varepsilon$, an adversary will observe almost the same output regardless of the presence (or value \rev{change}) of any one individual in the dataset; this is what provides privacy protection for that individual.  

\subsection{Private ADMM proposed in \cite{zhang2017}} 
Two randomizations were proposed in \cite{zhang2017}: (i) dual variable perturbation, where each node $i$ adds a random noise to its dual variable $\lambda_i(t)$ before updating its primal variable $f_i(t)$ using \eqref{eq:prelimi_8} in each iteration; and (ii) primal variable perturbation, where after updating primal variable $f_i(t)$, each node adds a random noise to it before broadcasting to its neighbors. Both were evaluated for a single iteration for a fixed privacy constraint. As we will see later in numerical experiments, \rev{the privacy loss accumulates significantly when inspected over multiple iterations.}

In contrast, in this study we will explore the use of the penalty \rev{parameter} $\eta$ to provide privacy.  In particular, we will allow this to be private information to every node, i.e., each decides its own $\eta$ in every iteration and it is not exchanged among the nodes. 
Below we will begin by modifying the ADMM to accommodate private penalty terms.  


\section{Modified ADMM (M-ADMM)} \label{sec:m-admm} 

\subsection{Making $\eta$ a node's private information} 

Conventional ADMM \cite{boyd2011} requires that the penalty parameter $\eta$ be fixed and equal to the dual updating step size for all nodes in all iterations. Varying the penalty parameter to accelerate convergence in ADMM has been proposed in the literature. For instance, \cite{he2002,magnusson2014,aybat2015,xu2016} vary this penalty parameter in every iteration but keep it the same for different equality constraints in \eqref{eq:prelimi_2}. In \cite{song2016,zhang2017privacy} this parameter varies in each iteration and is allowed to differ for different equality constraints. However, all of these modifications are based on the original ADMM (Eqn. \eqref{eq:prelimi_4}-\eqref{eq:prelimi_7}) and not on the simplified version (Eqn. \eqref{eq:prelimi_8}-\eqref{eq:prelimi_9}); the significance of this difference is discussed below in the context of privacy requirement. Moreover, we will decouple $\eta_i(t+1)$ from the dual updating step size, denoted as $\theta$ below. For simplicity, $\theta$ is fixed for all nodes in our analysis, but can also be private information as we show in numerical experiments. 


First consider replacing $\eta$ with $\eta_{ij}(t+1)$ in Eqn. \eqref{eq:prelimi_4}-\eqref{eq:prelimi_5} of the original ADMM (as is done in \cite{song2016,zhang2017privacy}) and replacing $\eta$ with $\theta$ in Eqn. \eqref{eq:prelimi_6}-\eqref{eq:prelimi_7}; we obtain the following: 
\begin{eqnarray}
f_i(t+1) = \underset{f_i}{\text{argmin}}\ \{O(f_i,D_i)  + 2\lambda_i(t)^Tf_i \nonumber\\ + \sum_{j \in \mathscr{V}_i}\frac{\eta_{ij}(t+1)+\eta_{ji}(t+1) }{2}||\dfrac{1}{2}(f_i(t)+f_j(t))-f_i||_2^2\}~;\nonumber\\
\lambda_{i}(t+1) = \lambda_{i}(t) +  \dfrac{\theta}{2}\sum_{j \in \mathscr{V}_i}(f_i(t+1)-f_j(t+1))~.\nonumber 
\end{eqnarray}
This however violates our requirement that $\eta_{ji}(t)$ be node $j$'s private information since this is needed by node $i$ to perform the above computation.   
%
To resolve this, we instead start from the simplified ADMM, modifying Eqn. \eqref{eq:prelimi_8}-\eqref{eq:prelimi_9}: 
\begin{eqnarray}
f_i(t+1) = \underset{f_i}{\text{argmin}}\ \{O(f_i,D_i) + 2\lambda_i(t)^Tf_i \nonumber \\ 
+ \eta_i(t+1) \sum_{j \in \mathscr{V}_i}||f_i-\frac{1}{2}(f_i(t)+f_j(t))||_2^2~ \}~; 
\label{eq:modify_1} \\ 
\lambda_{i}(t+1) = \lambda_{i}(t) +  \frac{\theta}{2}\sum_{j \in \mathscr{V}_i}(f_i(t+1)-f_j(t+1))~, \label{eq:modify_2}
\end{eqnarray}
where $\eta_i(t+1)$ is now node $i$'s private information. Indeed $\eta_i(t+1)$ is no longer purely a penalty parameter related to any equality constraint in the original sense. We will however refer to it as the private penalty parameter for simplicity. The above constitutes the M-ADMM algorithm. 

\subsection{Convergence Analysis}

We next show that the M-ADMM (Eqn. \eqref{eq:modify_1}-\eqref{eq:modify_2})  converges to the optimal solution under a set of common technical assumptions.  Our proof is based on the method given in \cite{ling2016}.  

\textbf{\textit{Assumption 1}:} Function $O(f_i,D_i)$ is convex and continuously differentiable in $f_i$, $\forall i$.

\textbf{\textit{Assumption 2}:} The solution set to the original ERM problem \eqref{eq:prelimi_1} is nonempty and there exists at least one bounded element. 

The KKT optimality condition of the primal update \eqref{eq:modify_1} is:
\begin{eqnarray}\label{eq:modify_4}
0 = \nabla O(f_i(t+1),D_i) + 2\lambda_i(t)\nonumber\\ + \eta_i(t+1)\sum_{j \in \mathscr{V}_i}(2{f}_i(t+1)-({f}_i(t)+{f}_j(t)))~.
\end{eqnarray} 
We next rewrite \eqref{eq:modify_2}-\eqref{eq:modify_4} in matrix form. Define the adjacency matrix of the network $A\in \mathbb{R}^{N \times N}$ as
$$a_{ij} = \begin{cases}
1, \ \ \text{ if node } i\text{ and node }j \text{ are connected } \\
0, \ \ \text{ otherwise }~. 
\end{cases}$$
Stack the variables $f_i(t)$, $\lambda_i(t)$ and $\nabla O(f_i(t),D_i)$ for $i \in \mathscr{N}$ into matrices, i.e.,
$$ \hat{f}(t) = \begin{bmatrix}
f_1(t)^T\\
f_2(t)^T\\
\vdots\\
f_N(t)^T
\end{bmatrix}\in \mathbb{R}^{N\times d} \text{ , \ \   }\Lambda(t) = \begin{bmatrix}
\lambda_1(t)^T\\
\lambda_2(t)^T\\
\vdots\\
\lambda_N(t)^T
\end{bmatrix}\in \mathbb{R}^{N\times d} $$ 
$$ \nabla \hat{O}(\hat{f}(t),D_{all}) = \begin{bmatrix}
\nabla O(f_1(t),D_1)^T\\
\nabla O(f_2(t),D_2)^T\\
\vdots\\
\nabla O(f_N(t),D_N)^T
\end{bmatrix}\in \mathbb{R}^{N\times d} $$
Let $V_i = | \mathscr{V}_i|$ be the number of neighbors of node $i$, and define the degree matrix $D = \textbf{diag}([V_1;V_2;\cdots;V_N])\in \mathbb{R}^{N \times N}$.
Define for the $t$-th iteration a penalty-weighted matrix $W(t) = \textbf{diag}([{\eta}_1(t);{\eta}_2(t);\cdots;{\eta}_N(t)])\in \mathbb{R}^{N \times N}$. 
Then the matrix form of \eqref{eq:modify_2}-\eqref{eq:modify_4} are:
\begin{eqnarray}
\nabla\hat{O}(\hat{f}(t+1),D_{all}) + 2\Lambda(t)+2W(t+1)D\hat{f}(t+1) \nonumber\\ 
- W(t+1)(D+A)\hat{f}(t)=\textbf{0}_{N\times d}~; 
\label{eq:converge_5} \\ 
2\Lambda(t+1) = 2\Lambda(t) +\theta (D-A)\hat{f}(t+1)~. 
\label{eq:converge_6}
\end{eqnarray}
Note that $D-A$ is the Laplacian matrix and $D+A$ is the signless Laplacian matrix of the network, with the following properties if the network is connected: {(i)} $D\pm A \succeq 0$ is positive semi-definite; {(ii)} $\text{Null}(D-A) = c\textbf{1}$, i.e., every member in the null space of $D-A$ is a scalar multiple of \textbf{1} with \textbf{1} being the vector of all $1$'s \cite{Jonathan2007}. 

Let $\sqrt{X}$ denote the square root of a symmetric positive semi-definite (PSD) matrix $X$ that is also symmetric PSD, i.e., $\sqrt{X}\sqrt{X} = X$. Define matrix $Y(t)$ such that $2\Lambda(t)=\sqrt{D-A} Y(t)$. Since $\Lambda(0) = \textbf{zeros}(N,d)$, which is in the column space of $D-A$, this together with \eqref{eq:converge_6} imply that $\Lambda(t)$ is in the column space of $D-A$ and $\sqrt{D-A}$.  This guarantees the existence of $Y(t)$. This allows us to rewrite \eqref{eq:converge_5}-\eqref{eq:converge_6} as:
\begin{eqnarray}
\nabla \hat{O}(\hat{f}(t+1),D_{all}) + \sqrt{D-A} Y(t+1)\nonumber\\
+(W(t+1)-\theta I )(D-A)\hat{f}(t+1)\nonumber\\ 
+ W(t+1)(D+A)(\hat{f}(t+1)-\hat{f}(t))=\textbf{0}_{N\times d}~; 
\label{eq:converge_7}\\%
Y(t+1) =  Y(t)+\theta \sqrt{D-A}\hat{f}(t+1)~. 
\label{eq:converge_8}
\end{eqnarray}

\begin{lemma}\label{Lemma:1}
[\textit{First-order Optimality Condition} \cite{ling2016}] Under Assumptions 1 and 2, the following two statements are equivalent:
\begin{itemize}
\item $\hat{f}^* = [(f_1^*)^T;(f_2^*)^T;\cdots;(f_N^*)^T] \in \mathbb{R}^{N\times d}$ is consensual, i.e., $f_1^*=f_2^*=\cdots=f_N^*=f_c^*$ where $f_c^*$ is the optimal solution to \eqref{eq:prelimi_1}.
\item There exists a pair $(\hat{f}^*,Y^*)$ with $Y^* = \sqrt{D-A} X$ for some $X\in \mathbb{R}^{N\times d}$ such that
\begin{eqnarray}
\nabla \hat{O}(\hat{f}^*,D_{all})+\sqrt{D-A} Y^*=\textbf{0}_{N\times d} ~; 
\label{eq:converge_9}\\
\sqrt{D-A}\hat{f}^* = \textbf{0}_{N\times d}~. 
\label{eq:converge_10}
\end{eqnarray}
\end{itemize}
\end{lemma}
Lemma \ref{Lemma:1} shows that a pair $(Y^*,\hat{f}^*)$ satisfying \eqref{eq:converge_9}\eqref{eq:converge_10} is equivalent to the optimal solution of our problem, hence the convergence of M-ADMM is proved by showing that $(Y(t),\hat{f}(t))$ converges to a pair $(Y^*,\hat{f}^*)$ satisfying \eqref{eq:converge_9}\eqref{eq:converge_10}. 

\begin{theorem}\label{Theorem:1}
Consider the modified ADMM defined by \eqref{eq:modify_1}-\eqref{eq:modify_2}. Let $\{Y(t),\hat{f}(t)\}$ be outputs in each iteration and $(Y^*,\hat{f}^*)$ a pair satisfying \eqref{eq:converge_9}-\eqref{eq:converge_10}. 
Denote $$Z(t) = \begin{bmatrix}
Y(t)\\
\hat{f}(t)
\end{bmatrix} \in \mathbb{R}^{2N \times d}, \ \ \ Z^* = \begin{bmatrix}
Y^*\\
\hat{f}^*
\end{bmatrix}\in \mathbb{R}^{2N \times d}$$ 
$$ J(t) = \begin{bmatrix}
\frac{I_{N \times N}}{\theta} & 0\\
0 & W(t)(D+A)
\end{bmatrix}\in \mathbb{R}^{2N \times 2N}$$
Let $\langle \cdot,\cdot \rangle_F$ be the Frobenius inner product of two matrices. We have
\begin{equation}\label{eq:converge_13}
\langle Z(t+1)-Z^*, J(t+1)(Z(t+1)-Z(t))\rangle_F \leq 0~. 
\end{equation}
If ${\eta}_i(t+1)\geq{\eta}_i(t) \geq \theta > 0$ and ${\eta}_i(t)<+\infty$, $\forall t,i$,  then $(Y(t),\hat{f}(t))$ converges to $(Y^*,\hat{f}^*)$.
\end{theorem}

\subsection{Convergence Rate Analysis}
To further establish the convergence rate of modified ADMM, an additional assumption is used:

\textbf{ \textit{Assumption 3:}} For all $i \in \mathscr{N}$, $O(f_i,D_i)$ is strongly convex in $f_i$ and has Lipschitz continues gradients, i.e., for any $f_i^1$ and ${f}_i^2$, we have:  
\begin{equation}\label{eq:converge_1}
(f_i^1 - f_i^2)^T(\nabla O(f_i^1,D_i)-\nabla O(f_i^2,D_i))\geq m_i||f_i^1- f_i^2||_2^2 \nonumber \\
\end{equation} 
\begin{equation} 
||\nabla O(f_i^1,D_i)-\nabla O(f_i^2,D_i)||_2 \leq M_i||f_i^1- f_i^2||_2  
\label{eq:converge_2}
\end{equation}
where $m_{i}>0$ is the strong convexity constant and $0 <M_{i}<+\infty$ is the Lipschitz constant.


\begin{theorem}\label{Theorem:2}
Define $D_m = \textbf{diag}([m_1;m_2;\cdots;m_N]) \in \mathbb{R}^{N \times N}$ and $D_M = \textbf{diag}([M_1^2;M_2^2;\cdots;M_N^2])\in \mathbb{R}^{N \times N}$ with $m_i>0$ and $0<M_i<+\infty$ as given in Assumption 3. 
%
Denote by $||X||^2_{J} = \langle X, JX \rangle_F$ the Frobenius inner product of any matrix $X$ and $JX$; denote by $\sigma_{\text{min}}(\cdot)$ and $\sigma_{\text{max}}(\cdot)$ the smallest nonzero, and the largest, singular values of a matrix, respectively.

Let $\tilde{\sigma}_{\text{max}}(t)=\sigma_{\text{max}}({W(t)(D+A)})$, $\bar{\sigma}_{\text{max/min}}(t) = \sigma_{\text{max/min}}({(W(t)-\theta I )(D-A)})$ and $\mu>1$ be an arbitrary constant. Consider any $\delta(t)$ that satisfies \eqref{eq: L4_1}\eqref{eq: L4_2}: 
\begin{equation}\label{eq: L4_1}
\frac{\delta(t)\mu^2\tilde{\sigma}_{\text{max}}(t)}{\theta\sigma_{\text{min}}(D-A)}\leq 1
\end{equation}
and 
\begin{equation}\label{eq: L4_2}
\begin{gathered}
\delta(t)(\frac{\mu\bar{\sigma}_{\text{max}}(t)^2\textbf{I}_N+\mu^2D_M}{\theta\sigma_{\text{min}}(D-A)(\mu-1)}+W(t)(D+A))\\ \preceq  2(W(t)-\theta I )(D-A) + 2D_m~. 
\end{gathered}
\end{equation}
If $\eta_i(t+1)\geq \eta_i(t)\geq \theta > 0$ and $\eta_i(t)<+\infty$, $\forall t,i$, then $(Y(t),\hat{f}(t))$ converges to $(Y^*,\hat{f}^*)$ in the following sense:
 \begin{equation*}\label{eq:T2_1}
(1+\delta(t))||Z(t)-Z^*||^2_{J(t)} \leq ||Z(t-1)-Z^*||^2_{J(t)}~. 
\end{equation*}
Furthermore, a lower bound on $\delta(t)$ is:
\begin{equation}\label{eq:T2_2}
\begin{gathered}
\min \{\frac{\theta\sigma_{\text{min}}(D-A)}{\mu^2\tilde{\sigma}_{\text{max}}(t)}, \frac{2m_o + 2\bar{\sigma}_{\text{min}}(t)}{\frac{\mu^2M_O^2+\mu\bar{\sigma}_{\text{max}}(t)^2}{\theta\sigma_{\text{min}}(D-A)(\mu-1)} + \tilde{\sigma}_{\text{max}}(t)}\}
\end{gathered}
\end{equation}
where $m_o = \min_{i \in \mathscr{N}}\{m_i\}$ and $M_O = \max_{i \in \mathscr{N}}\{M_i\}$.
\end{theorem}

Although Theorem \ref{Theorem:2} only gives a lower bound on the convergence rate ($1+\delta(t)$) of the M-ADMM, it reflects the impact of penalty $\{\eta_i(t)\}_{i=1}^N$ on the convergence. 
Since $\bar{\sigma}_{\text{max}}(t) = \sigma_{\text{max}}({(W(t)-\theta I )(D-A)})$ and $\tilde{\sigma}_{\text{max}}(t)=\sigma_{\text{max}}({W(t)(D+A)})$, larger penalty 
results in larger $\bar{\sigma}_{\text{max}}(t)$ and $\tilde{\sigma}_{\text{max}}(t)$. By \eqref{eq:T2_2}, the first term, $\frac{\theta\sigma_{\text{min}}(D-A)}{\mu^2\tilde{\sigma}_{\text{max}}(t)}$ is smaller when $\tilde{\sigma}_{\text{max}}(t)$ is larger.  The second term is bounded by $\frac{\theta\sigma_{\text{min}}(D-A)(\mu-1)(2m_o+2\bar{\sigma}_{\text{min}}(t))}{\mu\bar{\sigma}_{\text{max}}(t)^2}$, which is smaller when $\bar{\sigma}_{\text{max}}(t)$ is larger. Therefore, the convergence rate $1+\delta(t)$ decreases as $\{\eta_i(t)\}_{i=1}^N$ increase.


\section{Private M-ADMM}\label{sec:private} 

In this section we present a privacy preserving version of M-ADMM.  
To begin, a random noise $\epsilon_i(t+1)$ with probability density proportional to $\exp\{-\alpha_i(t+1)||\epsilon_i(t+1)||_2\}$  is added to penalty term in the objective function of \eqref{eq:modify_1}:
\begin{equation}\label{eq:P_modify_1}
\begin{gathered}
{L}_i^{priv} (t+1) = O(f_i,D_i)  + 2\lambda_i(t)^Tf_i \\ + \eta_i(t+1) \sum_{j \in \mathscr{V}_i}||f_i+\epsilon_i(t+1)-\frac{1}{2}(f_i(t)+f_j(t))||_2^2
\end{gathered}
\end{equation}

To generate this noisy vector, choose the norm from the gamma distribution \rev{with shape $d$ and scale $\frac{1}{\alpha_i(t+1)}$} and the direction uniformly, where $d$ is the dimension of the feature space. Then node $i$'s local result is obtained by finding the optimal solution to the private objective function: 
\begin{equation}\label{eq:P_modify_2}
f_i(t+1) = \underset{f_i}{\text{argmin}}\ {L}_i^{priv}(t+1) , \ \ i \in \mathscr{N}~. 
\end{equation}
It is equivalent to \eqref{eq:P_modify_3} below when noise $\eta_i(t+1)V_i\epsilon_i(t+1)$ is added to the dual variable $\lambda_i(t)$:  
\begin{equation}\label{eq:P_modify_3}
\begin{gathered}
\underset{f_i}{\text{argmin}}\  \tilde{L}_i^{priv} (t+1) = \frac{C}{B_i}\sum_{n=1}^{B_i} {\mathscr{L}}(y_{i}^n f_i^Tx_{i}^n ) + \frac{\rho}{N} R(f_i)\\  + 2(\lambda_i(t) + \eta_i(t+1)V_i\epsilon_i(t+1))^Tf_i \\+\eta_i(t+1) \sum_{j \in \mathscr{V}_i}||f_i-\dfrac{1}{2}(f_i(t)+f_j(t))||_2^2~. \nonumber 
\end{gathered}
\end{equation}
Further, if $\eta_i(t+1)=\eta=\theta, \forall i, t$, then the above is reduced to the dual variable perturbation in \cite{zhang2017}\footnote{Only a single iteration is considered in \cite{zhang2017} while imposing a privacy constraint. Since we consider the entire iterative process, we don't impose per-iteration privacy constraint but calculate the total privacy loss.}.


The complete procedure is shown in Algorithm 1, where the condition used to generate $\theta$ helps bound the worst-case privacy loss but is not necessary in guaranteeing convergence.
\begin{algorithm}[tb]
   \caption{Penalty perturbation (PP) method}
   \label{alg1}
\begin{algorithmic}
 \STATE {\bfseries Parameter:} Determine $\theta$ such that $2c_1<\frac{B_i}{C}(\frac{\rho}{N} + 2\theta V_i)$ holds for all $i$.
  \STATE {\bfseries Initialize:} Generate $f_i(0)$ randomly and $\lambda_i(0) = \textbf{0}_{d \times 1}$ for every node $i \in \mathscr{N}$, $t=0$
   \STATE {\bfseries Input:} $\{D_i\}_{i=1}^N$, $\{\alpha_i(1),\cdots, \alpha_i(T)\}_{i=1}^N$ 
    \FOR{$t=0$ {\bfseries to} $T-1$}
    \FOR{$i=1$ {\bfseries to} $N$}
    \STATE  Generate noise $\epsilon_i(t+1) \sim \exp(-\alpha_i(t+1)||\epsilon||_2)$
     
    \STATE Perturb the penalty term according to \eqref{eq:P_modify_1}
     
    \STATE Update primal variable via \eqref{eq:P_modify_2}
    \ENDFOR
     \FOR{$i=1$ {\bfseries to} $N$}
    \STATE Broadcast $f_i(t+1)$ to all neighbors $j \in \mathscr{V}_i$
     \ENDFOR
     \FOR{$i=1$ {\bfseries to} $N$}
     \STATE Update dual variable according to \eqref{eq:modify_2}
     \ENDFOR 
     \ENDFOR
    \STATE {\bfseries Output:} upper bound of the total privacy loss $\beta$
\end{algorithmic}
\end{algorithm}

In a distributed and iterative setting, the ``output'' of the algorithm is not merely the end result, but includes all intermediate results generated and exchanged during the iterative process. For this reason, we formally state the differential privacy definition in this setting below. 
\begin{definition}\label{Def}
Consider a connected network $G(\mathscr{N},\mathscr{E})$ with a set of nodes $\mathscr{N} = \{1,2,\cdots,N\}$. Let $f(t) = \{f_i(t)\}_{i=1}^N$ denote the information exchange of all nodes in the $t$-th iteration.  
A distributed algorithm is said to satisfy $\beta$-differential privacy during $T$ iterations if for any two datasets $D_{all} = \cup_i D_i$ and $\hat{D}_{all} = \cup_i \hat{D}_i$, differing in at most one data point, and for any set of possible outputs $S$ during $T$ iterations, the following holds:
\begin{equation*}
\begin{gathered}
\frac{\text{Pr}(\{f(t)\}_{t=0}^T \in S|D_{all})}{\text{Pr}(\{f(t)\}_{t=0}^T \in S|\hat{D}_{all})} \leq \exp(\beta)
\end{gathered}
\end{equation*}
\end{definition}

We now state our main result on the privacy property of the penalty perturbation algorithm using the above definition. Additional assumptions on $\mathscr{L}(\cdot)$ and $R(\cdot)$ are used. 

\rev{\textbf{\textit{Assumption 4}:} The loss function $\mathscr{L}$ is strictly convex and twice differentiable. $|\mathscr{L}'| \leq 1$ and $0 <\mathscr{L}''\leq c_1$ with $c_1$ being a constant.}

\rev{\textbf{\textit{Assumption 5}:} The regularizer $R$ is $1$-strongly convex and twice continuously differentiable.}
\begin{theorem}\label{thmP}
Normalize feature vectors in the training set such that $||x_{i}^n||_2\leq 1$ for all $i \in \mathscr{N}$ and $n$. Then the private M-ADMM algorithm (PP) satisfies the $\beta$-differential privacy with 
\begin{equation}
\beta \geq \underset{i \in \mathscr{N}}{\max}\{\sum_{t=1}^{T}\frac{C(1.4c_1+\alpha_i(t))}{\eta_i(t)V_iB_i}\}~. 
\end{equation}
\end{theorem}
\begin{figure}[t]
	\centering   
	\subfigure[Different $\eta_i(t)$]{\label{fig1:a}\includegraphics[width=40.8mm]{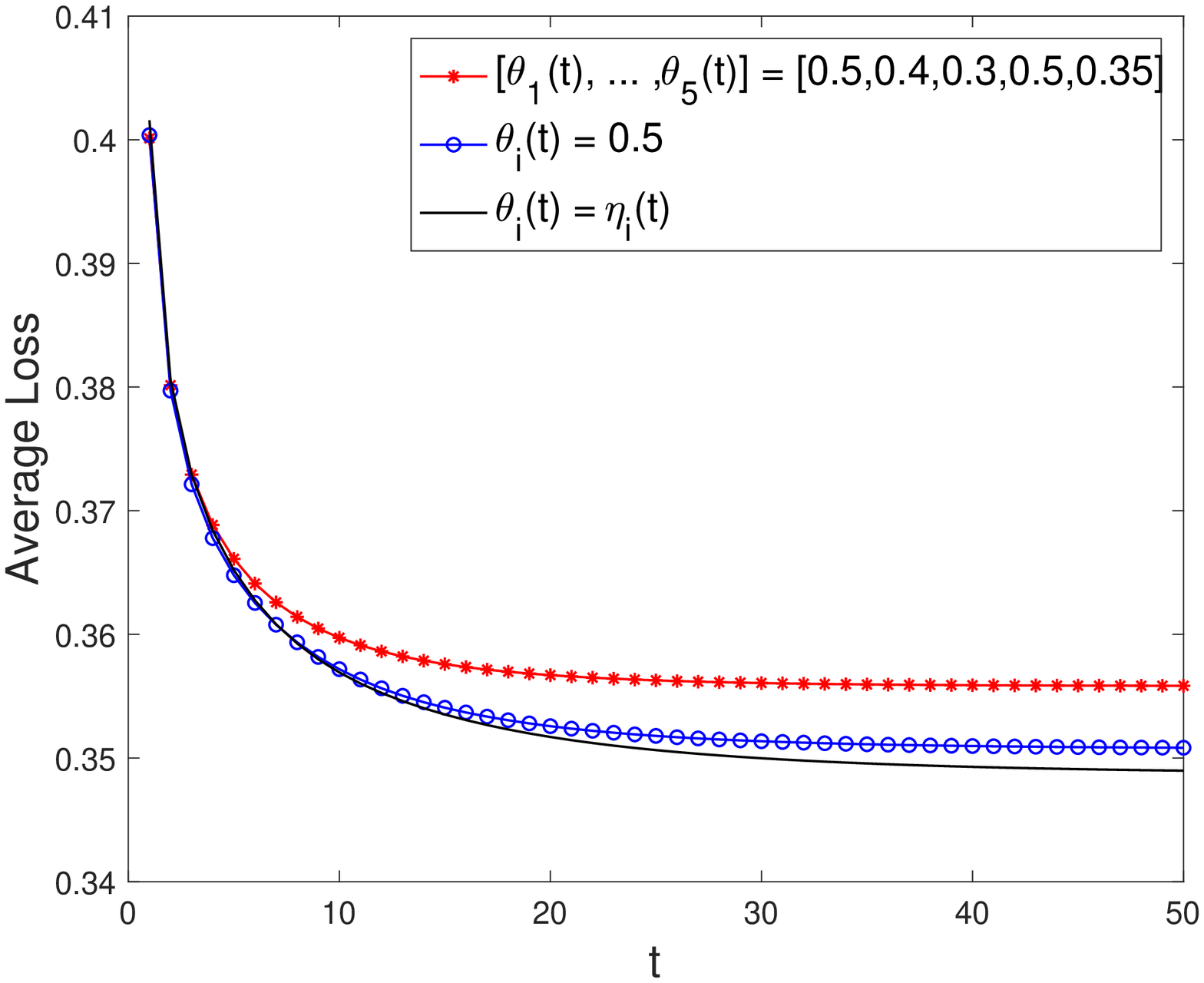}}
	\subfigure[Impact of increasing penalty]{\label{fig1:b}\includegraphics[width=40.8mm]{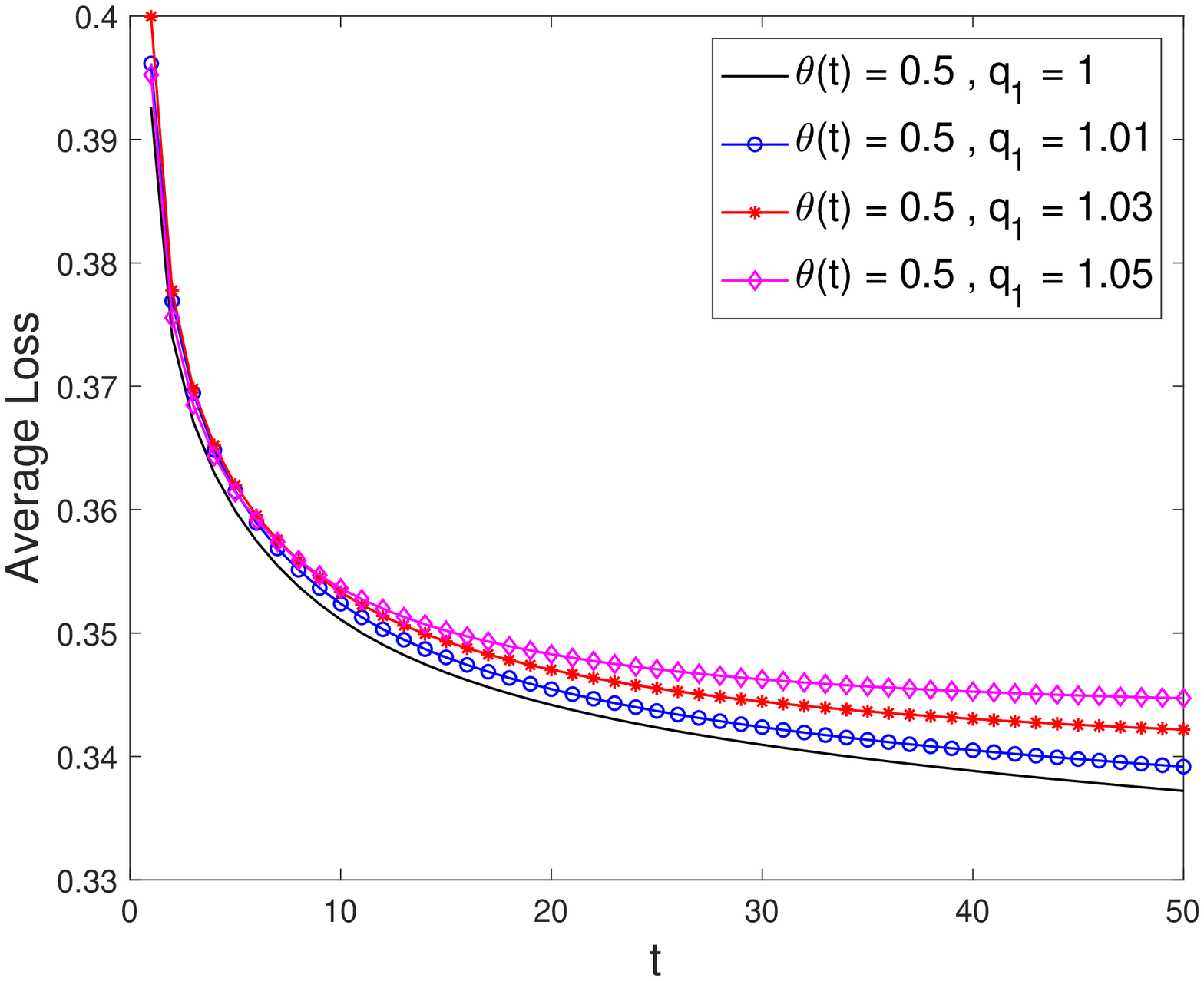}}
	\caption{Convergence properties of M-ADMM.}
\end{figure}
\begin{figure}[t]
	\centering   
	\subfigure[Accuracy comparison:  $\alpha(t)=3q_2^{t-1}$]{\label{fig2:a}\includegraphics[width=80mm]{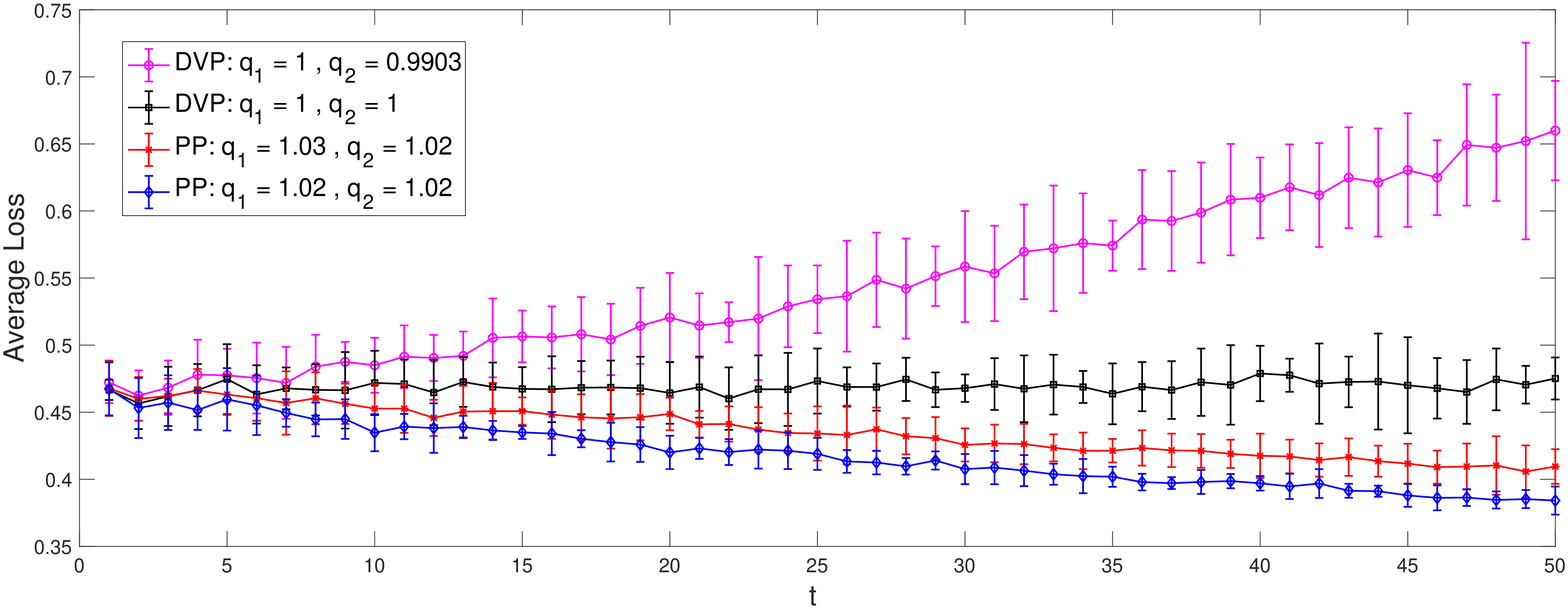}}
	\subfigure[Accuracy comparison: $\alpha(t)=5q_2^{t-1}$]{\label{fig2:b}\includegraphics[width=80mm]{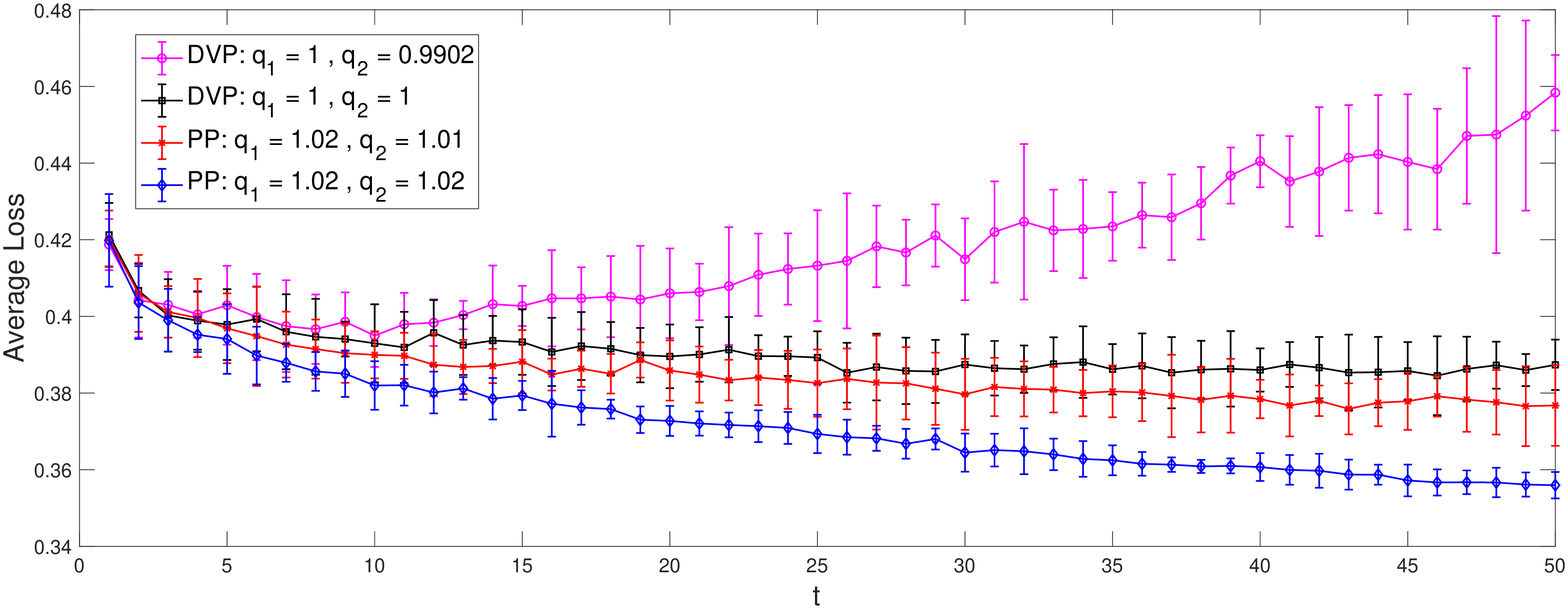}}
	\subfigure[Privacy:  $\alpha(t)=3q_2^{t-1}$]{\label{fig2:c}\includegraphics[width=37mm]{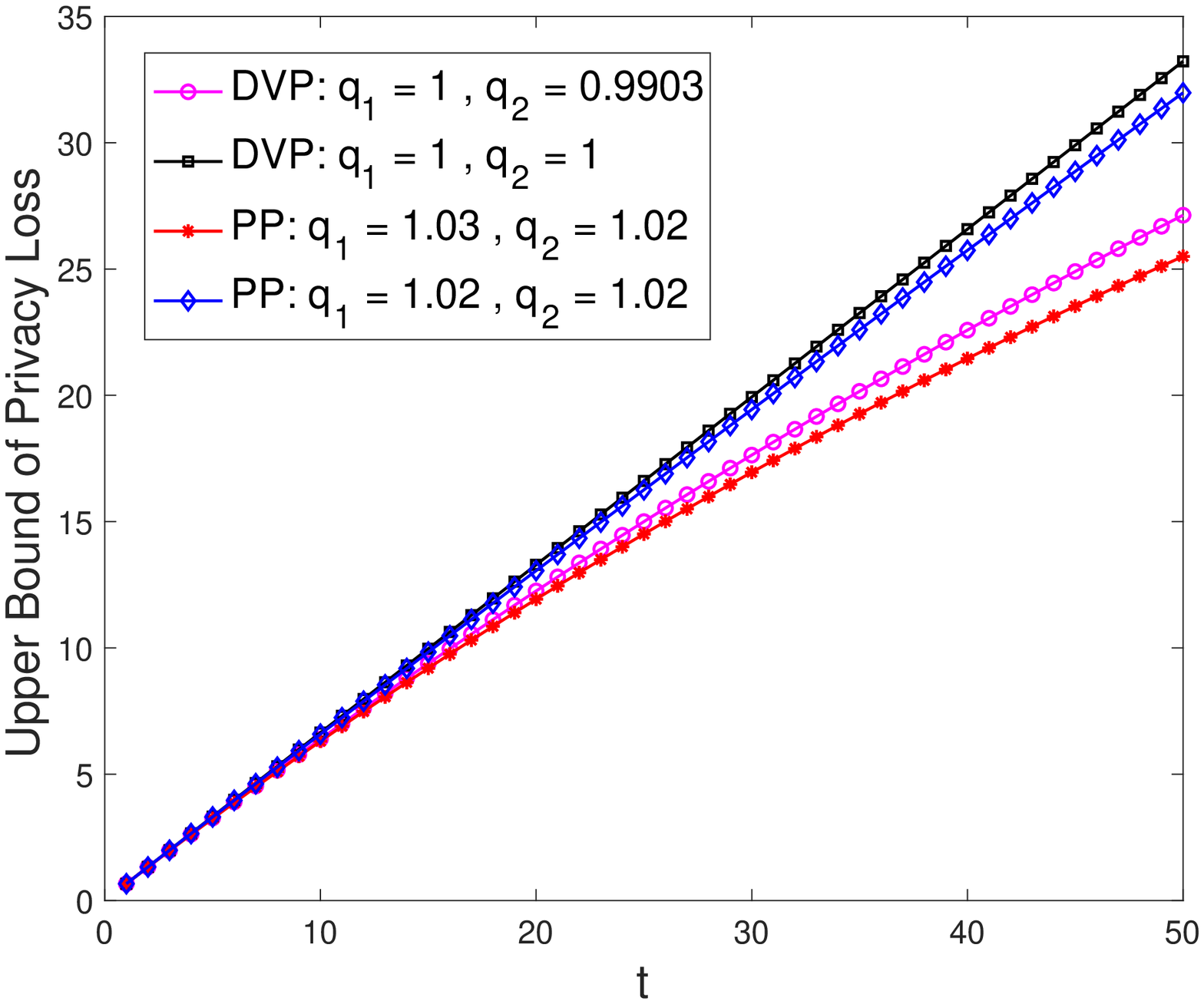}}
	\subfigure[Privacy:  $\alpha(t)=5q_2^{t-1}$]{\label{fig2:d}\includegraphics[width=37mm]{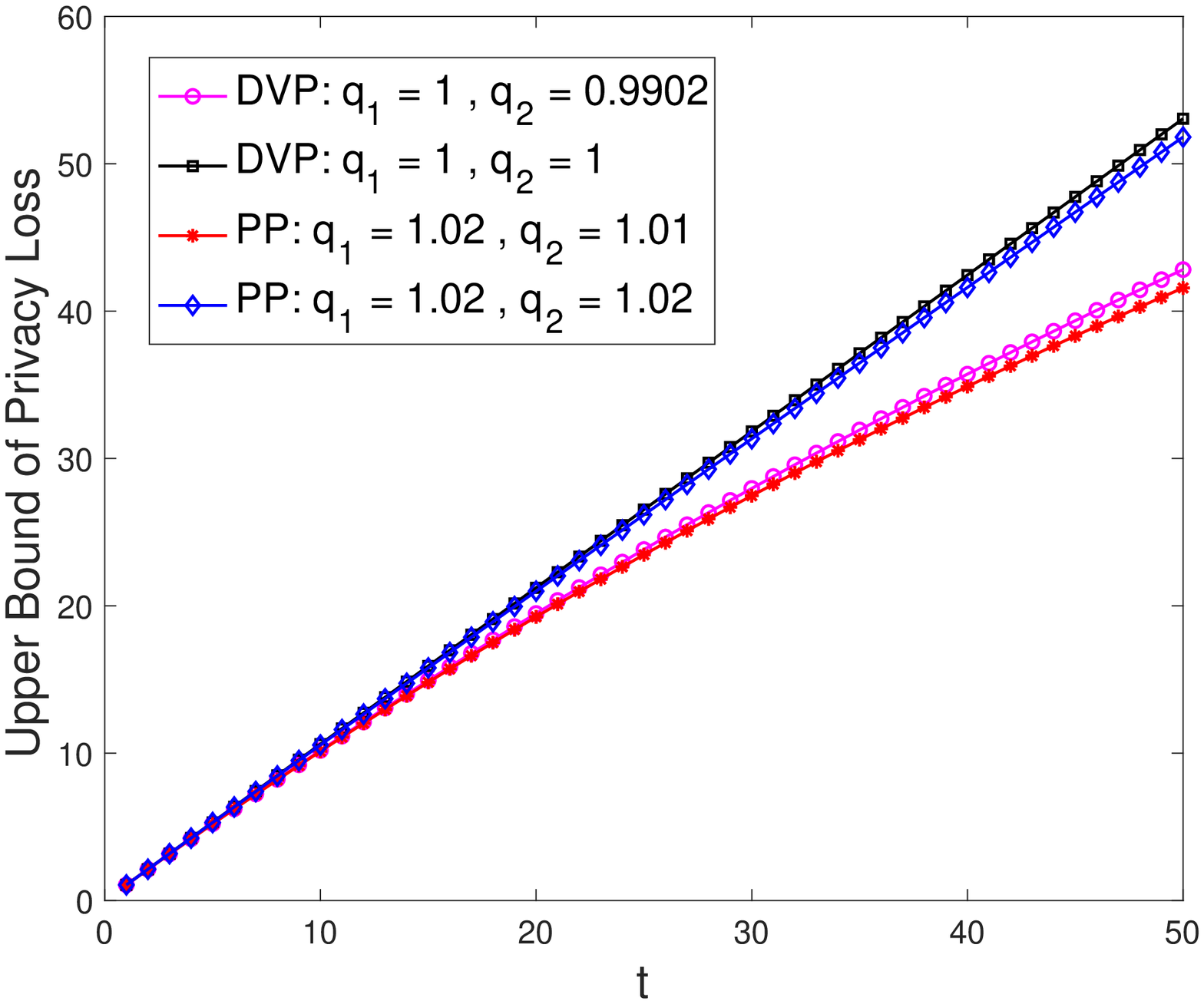}}
	\caption{Compare accuracy and privacy, $\eta(t)=0.5q_1^{t-1}$}
	\label{fig2}
\end{figure}
\section{Numerical Experiments}\label{sec:numerical} 
We use the same dataset as \cite{zhang2017}, i.e., the \textit{Adult} dataset from the UCI Machine Learning Repository \cite{Lichman2013}. It consists of personal information of around 48,842 individuals, including age, sex, race, education, occupation, income, etc. The goal is to predict whether the annual income of an individual is above \$50,000. 

To preprocess the data, we (1) remove all individuals with missing values; 
(2) convert each categorical attribute (with $m$ categories) to a binary vector of length $m$;   
(3) normalize columns (features) such that the maximum value of each column is 1; 
(4) normalize rows (individuals) such that its $l_2$ norm is at most 1; and 
(5) convert labels $\{\geq 50\text{k},\leq 50\text{k}\}$ to $\{+1,-1\}$. 
After this preprocessing, the final data includes 45,223 individuals, each represented as a 105-dimensional vector of norm at most 1.  

We will use as loss function the logistic loss $\mathscr{L}(z) = \log(1+\exp(-z))$, with $|\mathscr{L}'|\leq 1 $ and $\mathscr{L}'' \leq c_1 = \frac{1}{4}$. 
The regularizer is $R(f_i) = \frac{1}{2}||f_i||_2^2$. 
We will measure the accuracy of the algorithm by the average loss $L(t):=\frac{1}{N} \sum_{i=1}^{N}\frac{1}{B_i}\sum_{n=1}^{B_i}\mathscr{L}(y^n_if_i(t)^Tx^n_i) $ over the training set. We will measure the privacy of the algorithm by the upper bound $P(t):=\underset{i \in \mathscr{N}}{\max}\{\sum_{r=1}^{t}\frac{C(1.4c_1+\alpha_i(r))}{\eta_i(r)V_iB_i}\}$. 
The smaller $L(t)$ and $P(t)$, the higher accuracy and stronger privacy guarantee.


\subsection{Convergence of M-ADMM}
\rev{We consider a five-node network and}
assign each node the following private penalty parameters: $\eta_i(t) = \eta_i(1)q_i^{t-1}$ for node $i$, where $[\eta_1(1), \cdots,\eta_5(1)]$ $=[0.55,0.65,0.6,0.55,0.6]$ and $[q_1, \cdots, q_5]=$ $[1.01,1.03,1.1,1.2,1.02]$. 

Figure \ref{fig1:a} shows the convergence of M-ADMM under these parameters while using a fixed dual updating step size $\theta=0.5$ across all nodes (blue curve). This is consistent with Theorem \ref{Theorem:1}. As mentioned earlier, this step size can also be non-fixed (black) and different (red) for different nodes. 
In Figure \ref{fig1:b} we let each node use the same penalty $\eta_i(t)=\eta(t)=0.5q_1^{t-1}$ and compare the results by increasing $q_1$, $q_1\geq 1$.  We see that increasing penalty slows down the convergence, and larger increase in $q_1$ slows it down even more, which is consistent with Theorem \ref{Theorem:2}.


\subsection{Private M-ADMM}
We next inspect the accuracy and privacy of the penalty perturbation (PP) based private M-ADMM (Algorithm \ref{alg1}) and compare it with the dual variable perturbation (DVP) method proposed in \cite{zhang2017}. In this set of experiments, for simplicity of presentation we shall fix $\theta = 0.5$, let $\eta_i(t) = \eta(t) = \theta q_1^{t-1}$, and noise $\alpha_i(t) = \alpha(t) = \alpha(1)q_2^{t-1}$ for all nodes. We observe similar results when $\eta_i(t)$ and $\alpha_i(t)$ vary from node to node.

For each parameter setting, we perform 10 independent runs of the algorithm, and record both the mean and the range of their accuracy.  Specifically, $L^l(t)$  denotes the average loss over the training dataset in the $t$-th iteration of the $l$-th experiment ($1\leq l \leq 10$).  The mean of average loss is then given by $L_{mean}(t) = \frac{1}{10}\sum_{l=1}^{10} L^l(t)$, and the range $L_{range}(t) = \underset{1\leq l \leq 10}{\max} L^l(t) - \underset{1\leq l \leq 10}{\min} L^l(t)$. 
The larger the range $L_{range}(t)$ the less stable the algorithm, i.e., under the same parameter setting, the difference in performances (convergence curves) of every two experiments is larger. 
Each parameter setting also has a corresponding upper bound on the privacy loss denoted by $P(t)$. 
Figures \ref{fig2:a}\ref{fig2:b} show both $L_{mean}(t)$ and $L_{range}(t)$ as vertical bars centered at $L_{mean}(t)$.  Their corresponding privacy upper bound is given in Figures \ref{fig2:c}\ref{fig2:d}. The pair \ref{fig2:a}-\ref{fig2:c} (resp. \ref{fig2:b}-\ref{fig2:d}) is for the same parameter setting.

Figure \ref{fig2} compares PP (blue \& red, with $\eta_i(t)$ increasing geometrically) with DVP (black \& magenta, with $\eta_i(t)=\theta$, $\forall i,t$).
We see that in both cases improved accuracy comes at the expense of higher privacy loss (from magenta to black under DVP, from red to blue under PP).  However, we also see that with suitable choices of $q_1$, $q_2$, PP can outperform DVP significantly both in accuracy and in privacy (e.g., red outperforms magenta in both accuracy and privacy, and blue outperforms black in both accuracy and privacy).

\begin{figure*}[t]
	\centering   
	\includegraphics[width=80mm]{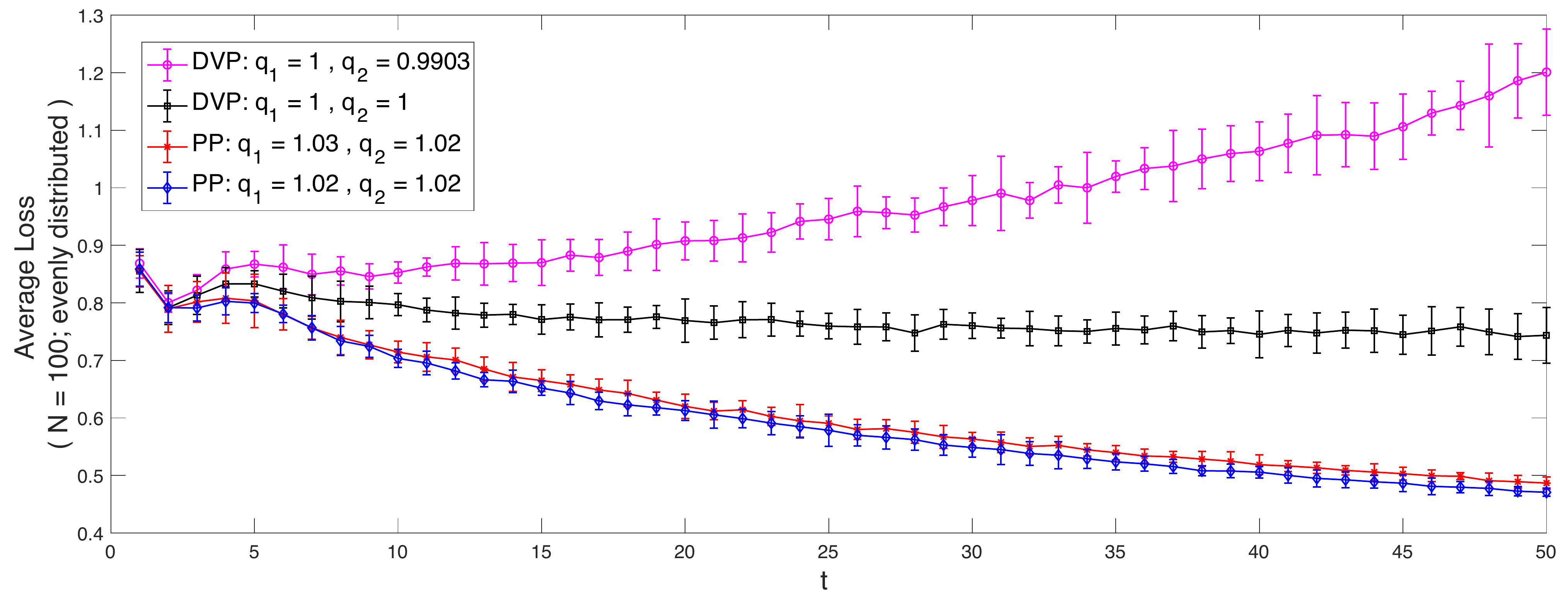}
	\includegraphics[width=37mm]{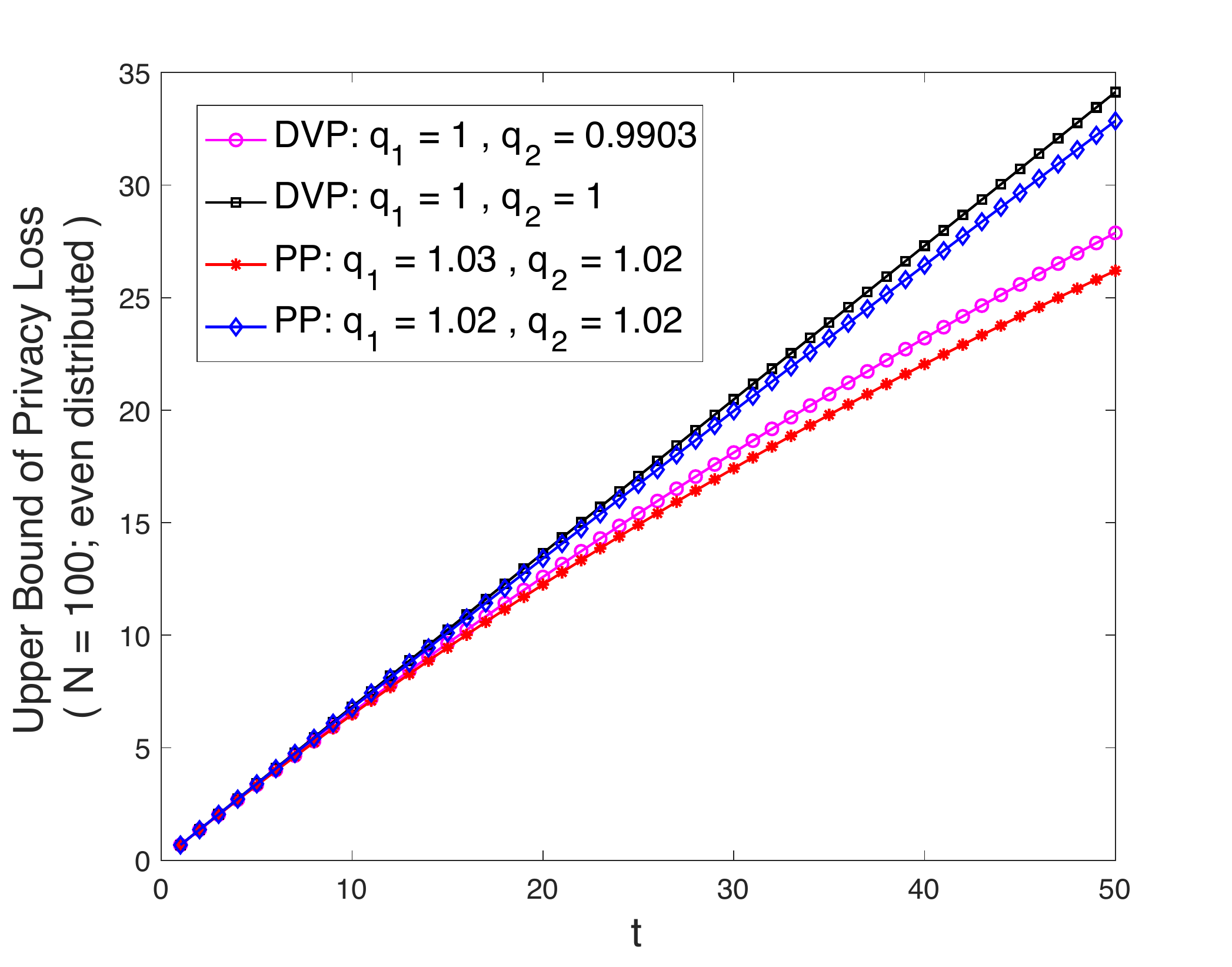}
	\includegraphics[width=37mm]{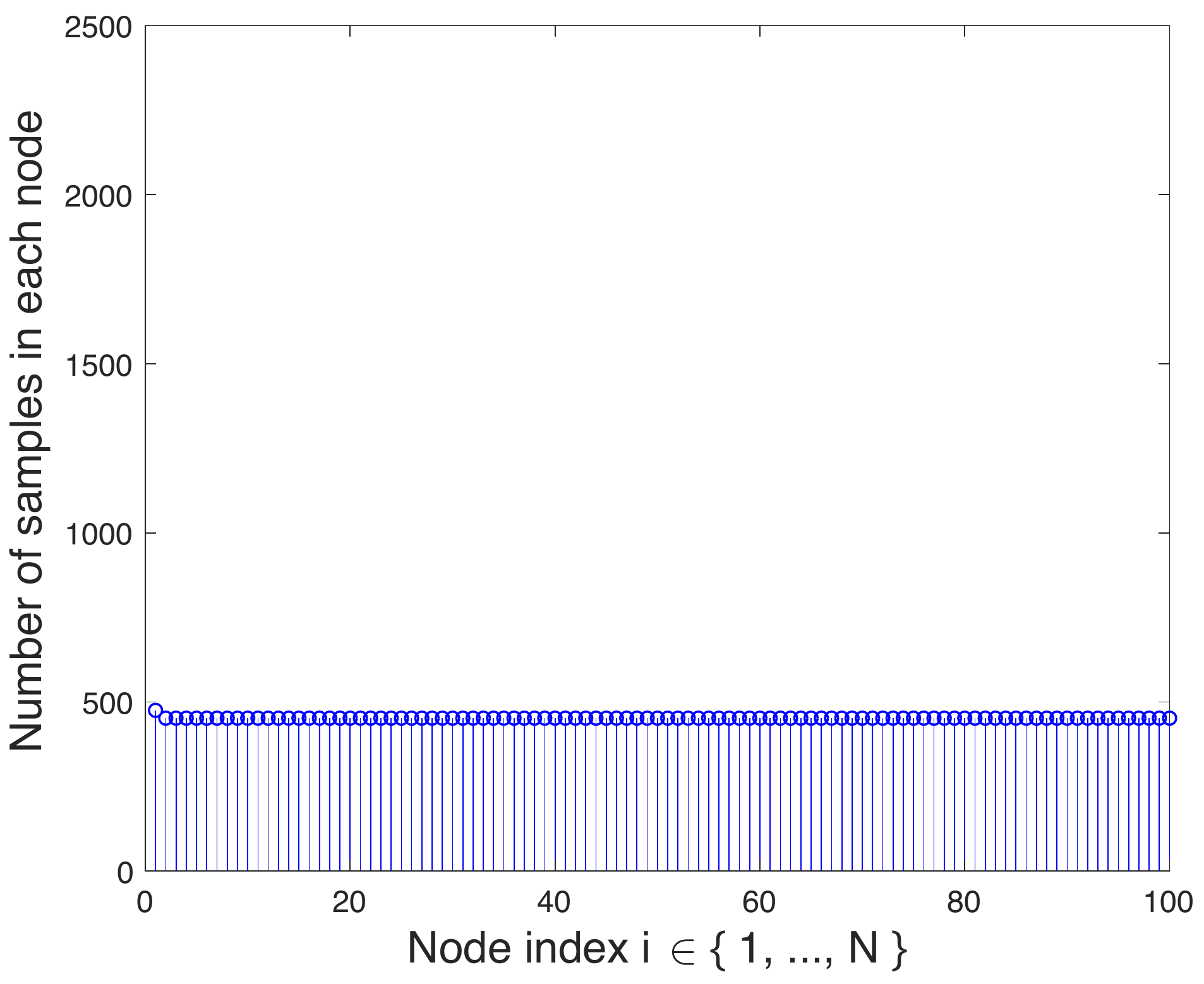}
	\caption{Compare accuracy and privacy for large network with data \textbf{evenly} spread across 100 nodes, $\eta(t)=0.5q_1^{t-1}$, $\alpha(t)=3q_2^{t-1}$}
	\label{fig3}
\end{figure*}

\begin{figure*}[t]
	\centering   
	\includegraphics[width=80mm]{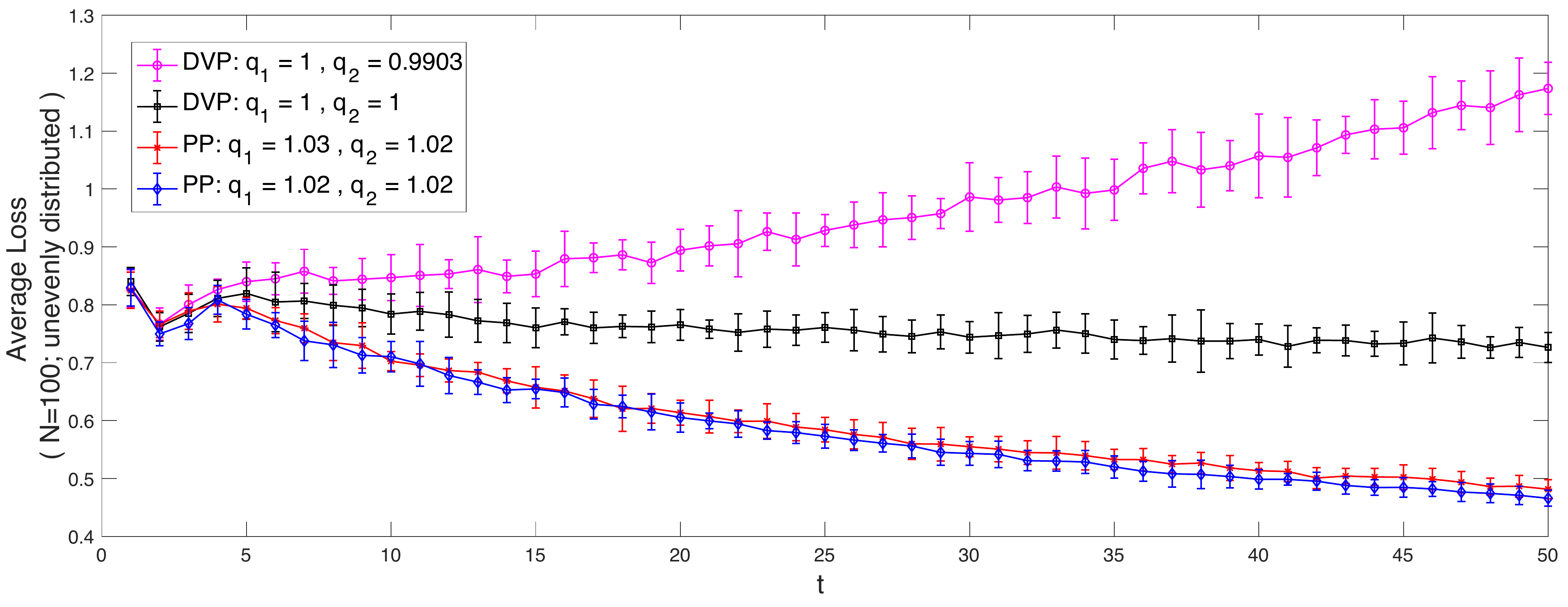}
	\includegraphics[width=37mm]{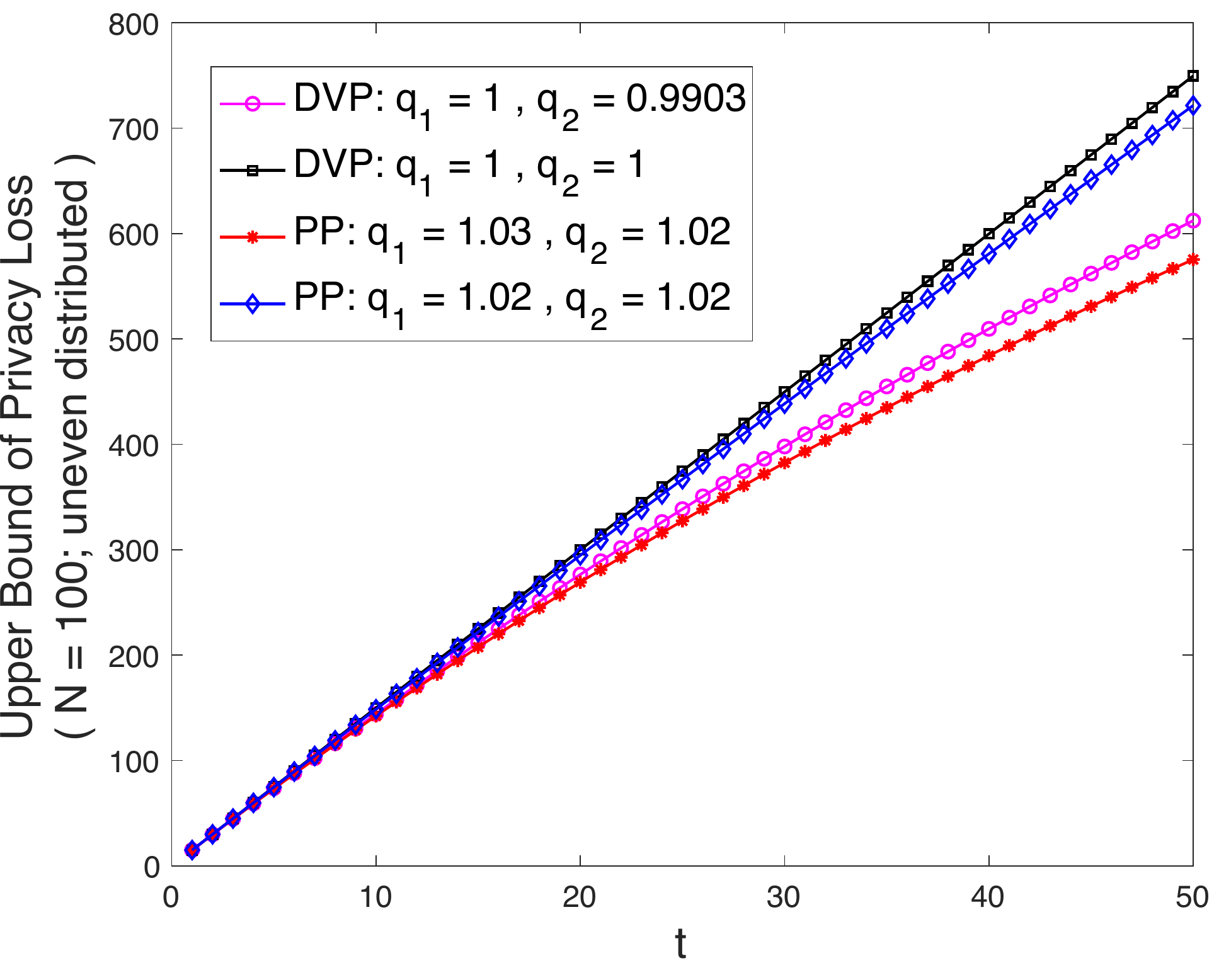}
	\includegraphics[width=37mm]{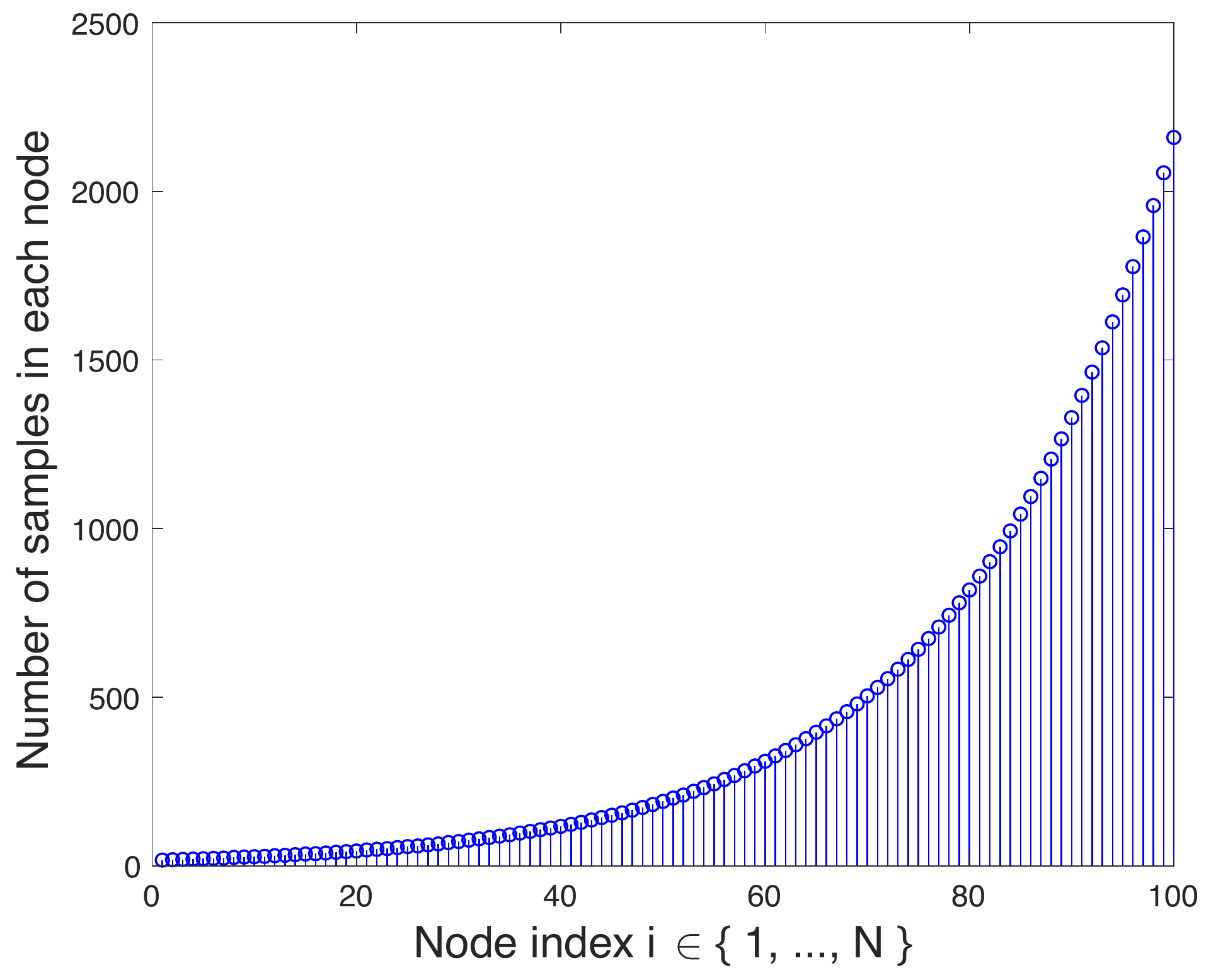}
	\caption{Compare accuracy and privacy for large network with data \textbf{unevenly} spread across 100 nodes, $\eta(t)=0.5q_1^{t-1}$, $\alpha(t)=3q_2^{t-1}$}
	\label{fig4}
\end{figure*}

\rev{We also performed experiments with the same dataset on larger networks with tens and hundreds of nodes and with samples evenly and unevenly spread across nodes. In both cases, convergence is attained and our algorithm continues to outperform \cite{zhang2017} in a large network (see Figures \ref{fig3} \& \ref{fig4}). Since the privacy loss of the network is dominated by the node with the largest privacy loss and it increases as the number of samples in a node decreases (Theorem \ref{thmP}), the loss of privacy in a network with uneven sample size distributions is higher; note that this is a common issue with this type of analysis.}

\section{Discussion}\label{sec:discuss} 
Our numerical results show that increasing the penalty $\{\eta_i(t)\}_{i=1}^N$ over iterations can improve the algorithm's accuracy and privacy simultaneously. Below we provide some insight on why this is the case and discuss possible generalizations of our method. 

\subsection{Higher accuracy}\label{subsec:accuracy} 
When the algorithm is perturbed by random noise, which is necessary to achieve privacy, increasing the penalty parameters over iterations makes the algorithm more noise resistant.  
In particular, for the minimization in \eqref{eq:P_modify_2}, larger $\eta_i(t+1)$ results in smaller updates of variables, i.e., smaller distance between $f_i(t+1)$ and $f_i(t)$.  In the non-private case, since $f_i(t)$ always moves toward the optimum, smaller update slows down the process.  In the private case, on the other hand, since a random noise is added to each update, $f_i(t)$ does not always move toward the optimum in each step. 
When the overall perturbation has a larger variance, it is more likely that $f_i(t)$ could move further away from the optimum in some iterations. Because larger $\eta_i(t)$ leads to smaller update, it helps prevent $f_i(t)$ from moving too far away from the optimum, thus stabilizing the algorithm (smaller $L_{range}(t)$). 

\subsection{Stronger privacy}
First of all, more added noise means stronger privacy guarantee. Increasing $\eta_i(t)$ and $\alpha_i(t)$ in such a way that the overall perturbation $2\eta_i(t)V_i\epsilon_i(t)^Tf_i(t)$ in \eqref{eq:P_modify_3} is increasing leads to less privacy loss, as shown in Figure \ref{fig2}. The noise resistance provided by an increasing $\eta_i(t)$ indeed allows larger noises to be added under PP without jeopardizing convergence as observed in Section \ref{subsec:accuracy}. 

More interestingly, keeping $\eta_i(t)$ private further strengthens privacy protection.
Consider the following threat model: An attacker knows $\{(x^n_i,y^n_i)\}_{n=2}^{B_i}$ and $\{f_j(t)\}_{j \in \mathscr{V}_i \cup i}$ for all $t$, i.e., all data points except for the first data point of node $i$, as well as all intermediate results of node $i$ and its neighbors. If the attacker also knows the dual updating step size $\theta$ and penalty parameter $\{\eta_i(t)\}_{t=1}^T$ of node $i$, it can then infer the unknown data point $(x_i^1,y_i^1)$ with high confidence \rev{by combining the KKT optimality conditions from all iterations (see supplementary material for details).} 
However, if the penalty parameters $\{\eta_i(t)\}_{t=1}^T$ are private to each node, then it is impossible for the attacker to infer \rev{the unknown data}. Even if the attacker knows the participation of an individual, it remains hard to infer its features.


\subsection{Generalization \& comparison}
\rev{The main contribution of this paper is the finding that increasing $\{\eta_i\}_{i=1}^N$ improves the algorithm's ability to resist noise: even though we increase noise in each iteration to improve privacy, the accuracy does not degrade significantly due to this increasing robustness, which improves the privacy-utility tradeoff. This property holds regardless of the noise distribution. While the present privacy analysis uses a similar framework as in \cite{chaudhuri2011,zhang2017} (objective perturbation with added Gamma noise), we can also use methods from other existing (centralized) ERM differentially private algorithms to every iteration in ADMM. For example, if we allow some probability ($\delta>0$) of violating $\epsilon$-differential privacy and adopt a weaker variant ($\epsilon$, $\delta$)-differential privacy, we can adopt methods from works such as \cite{kifer2012private,jain2014near,bassily2014differentially}, by adding Gaussian noise to achieve tighter bounds on privacy loss. However, as noted above, the robustness is improved as $\{\eta_i\}_{i=1}^N$ increases; thus the same conclusion can be reached that both privacy and accuracy can be improved. }

\rev{This idea can also be generalized to} other differentially private iterative algorithms. A key observation of our algorithm is that the overall perturbation ($2\eta_i(t)V_i\epsilon_i(t)^Tf_i(t)$) is related to the parameter that controls the updating step size ($\eta_i(t)$). 
In general, if the algorithm is perturbed in each iteration with a quantity $\phi(\epsilon,\xi)$, which is a function of added noise $\epsilon$ and some parameter $\xi$ that controls the step size, such that the resulting step size and $\phi(\epsilon,\xi)$ move in opposite directions (i.e., decreasing step size increases the $\phi(\epsilon,\xi)$), then it is possible to simultaneously improve both accuracy and privacy by varying $\xi$ to decrease the step size over time.

Interestingly, in a differentially private (sub)gradient-based distributed algorithm \cite{huang2015}, the step size and the overall perturbation move in the same direction (i.e., decreasing step size decreases perturbation). The reason for this difference is that under this subgradient-based algorithm, the sensitivity of the algorithm decreases with decreasing step size, which in turn leads to privacy constraint being satisfied with smaller perturbation. In contrast, for ADMM the sensitivity of the algorithm is independent of the step size, and the  perturbation actually needs to increase to improve privacy guarantee; the decreasing step size acts to compensate for this increase in noise to maintain accuracy, as discussed in Section \ref{subsec:accuracy}.

This issue of step size never arises in the study of \cite{zhang2017} because the analysis is only for a single iteration; however, as we have seen doing so leads to significant total privacy loss over many iterations.

\section{Conclusions}\label{sec:conclude} 
This paper presents a penalty-perturbation idea to introduce privacy preservation in iterative algorithms. 
We showed how to modify an ADMM-based distributed algorithm to improve privacy without compromising accuracy. The key idea is to add a perturbation correlated to the step size so that they change in opposite directions. Applying this idea to other iterative algorithms can be part of the future work.    


\section*{Acknowledgements}
This work is supported by the NSF under grants CNS-1422211, CNS-1646019, CNS-1739517.

\newpage 
\bibliography{icml18_Xueru}
\bibliographystyle{icml2018}


\onecolumn 
\icmltitle{Improving the Privacy and Accuracy of ADMM-Based Distributed Algorithms (Supplementary materials)}
\appendix
\section{Proof of Simplifying ADMM \cite{forero2010}}\label{appendix1}
By KKT condition of \eqref{eq:prelimi_5}, there is:
\begin{equation*}
0 = \lambda_{ij}^b(t) - \lambda_{ij}^a(t) + \eta(2 w_{ij}(t+1)-f_i(t+1)-f_j(t+1))
\end{equation*}

Implies:
\begin{equation}\label{eq:sim1}
w_{ij}(t+1) = \frac{1}{2\eta}( \lambda_{ij}^a(t) - \lambda_{ij}^b(t)) + \frac{1}{2}(f_i(t+1)+f_j(t+1))
\end{equation}

Plug \eqref{eq:sim1} into \eqref{eq:prelimi_6}\eqref{eq:prelimi_7}:
\begin{equation}\label{eq:sim2}
\lambda_{ij}^a(t+1) = \frac{1}{2}(\lambda_{ij}^a(t) + \lambda_{ij}^b(t)) + \frac{\eta}{2}(f_i(t+1)-f_j(t+1))
\end{equation}
\begin{equation}\label{eq:sim3}
\lambda_{ij}^b(t+1) = \frac{1}{2}(\lambda_{ij}^b(t) + \lambda_{ij}^a(t)) + \frac{\eta}{2}(f_i(t+1)-f_j(t+1))
\end{equation}

If initialize $\lambda_{ij}^a(0)=\lambda_{ij}^b(0)$ to be zero vectors for all node pairs $(i,j)$, \eqref{eq:sim2}\eqref{eq:sim3} imply that $\lambda_{ij}^a(t)=\lambda_{ij}^b(t)$ and $\lambda_{ji}^k(t) = -\lambda_{ij}^k(t), k \in \{a,b\}$ will hold for all $t$. \eqref{eq:sim1} becomes:
\begin{equation}\label{eq:sim4}
w_{ij}(t+1) =  \frac{1}{2}(f_i(t+1)+f_j(t+1))
\end{equation}

Let $\lambda_{ij}(t) = \lambda_{ij}^a(t)=\lambda_{ij}^b(t)$, \eqref{eq:prelimi_6}\eqref{eq:prelimi_7} can be simplified as:
\begin{equation}\label{eq:sim5}
\lambda_{ij}(t+1) = \lambda_{ij}(t) + \frac{\eta}{2}(f_i(t+1)-f_j(t+1))
\end{equation}

Plug \eqref{eq:sim4} into the augmented Lagrangian \eqref{eq:prelimi_3} to simplify it:
\begin{equation}\label{eq:sim6}
\begin{gathered}
L_\eta (\{f_i\},\{w_{ij},\lambda_{ij}^k\}) = \sum_{i=1}^N O(f_i,D_i) + \sum_{i=1}^N\sum_{j \in \mathscr{V}_i}(\lambda_{ij}(t))^T(f_i - f_j)\\ + \sum_{i=1}^N\sum_{j \in \mathscr{V}_i}\frac{\eta}{2}(||f_i-\frac{1}{2}(f_i(t)+f_j(t))||_2^2)  + \sum_{i=1}^N\sum_{j \in \mathscr{V}_i}\frac{\eta}{2}(||\frac{1}{2}(f_i(t)+f_j(t))-f_j||_2^2) 
\end{gathered}
\end{equation}

Since $\sum_{i=1}^N\sum_{j \in \mathscr{V}_i} \lambda_{ij}(t)f_j = \sum_{i=1}^N\sum_{j \in \mathscr{V}_i} \lambda_{ji}(t)f_i$ and $\lambda_{ij}(t) = -\lambda_{ji}(t)$, the second term in \eqref{eq:sim6} can be simplified:
\begin{equation*}
\sum_{i=1}^N\sum_{j \in \mathscr{V}_i}(\lambda_{ij}(t))^T(f_i - f_j) = 2\sum_{i=1}^N\sum_{j \in \mathscr{V}_i}(\lambda_{ij}(t))^Tf_i 
\end{equation*}

The last term can be expressed as:
\begin{equation*} 
\begin{gathered}
\sum_{i=1}^N\sum_{j \in \mathscr{V}_i}\frac{\eta}{2}(||\frac{1}{2}(f_i(t)+f_j(t))-f_j||_2^2)  = \sum_{i=1}^N\sum_{j \in \mathscr{V}_i}\frac{\eta}{2}(||\frac{1}{2}(f_i(t)+f_j(t))-f_i||_2^2) 
\end{gathered}
\end{equation*}

Therefore, \eqref{eq:sim6} is simplified as:
\begin{equation}\label{eq:sim7}
\begin{gathered}
L_\eta (\{f_i\},\{w_{ij},\lambda_{ij}^k\}) =\sum_{i=1}^N O(f_i,D_i) + 2\sum_{i=1}^N\sum_{j \in \mathscr{V}_i}\lambda_{ij}(t)^Tf_i + \sum_{i=1}^N\sum_{j \in \mathscr{V}_i}\eta(||f_i-\frac{1}{2}(f_i(t)+f_j(t))||_2^2)
\end{gathered}
\end{equation}

Define $\lambda_i(t) = \sum_{j \in \mathscr{V}_i}\lambda_{ij}(t)$. Based on \eqref{eq:sim5}\eqref{eq:sim7}, the original ADMM updates \eqref{eq:prelimi_4}-\eqref{eq:prelimi_7} are simplified as:

\begin{equation*}
\begin{gathered}
f_i(t+1) = \underset{f_i}{\text{argmin}}\ O(f_i,D_i) + 2\lambda_i(t)^Tf_i  + \eta \sum_{j \in \mathscr{V}_i}||f_i-\dfrac{1}{2}(f_i(t)+f_j(t))||_2^2
\end{gathered}
\end{equation*}
\begin{equation*}
\lambda_{i}(t+1) = \lambda_{i}(t) +  \dfrac{\eta}{2}\sum_{j \in \mathscr{V}_i}(f_i(t+1)-f_j(t+1))
\end{equation*}

\section{Proof of Theorem \ref{Theorem:1}}
Subtract \eqref{eq:converge_9} from \eqref{eq:converge_7} and \eqref{eq:converge_10} from \eqref{eq:converge_8}:
\begin{equation}\label{eq:theorem_1}
\begin{gathered}
\nabla  \hat{O}(\hat{f}(t+1),D_{all})-\nabla  \hat{O}(\hat{f}^*,D_{all})+ \sqrt{D-A} (Y(t+1)-Y^*)+(W(t+1)-\theta I )(D-A)\hat{f}(t+1)\\ + W(t+1)(D+A)(\hat{f}(t+1)-\hat{f}(t))=\textbf{0}_{N\times d}
\end{gathered}
\end{equation}
\begin{equation}\label{eq:theorem_2}
Y(t+1) =  Y(t)+\theta \sqrt{D-A}(\hat{f}(t+1)-\hat{f}^*)
\end{equation}

By convexity of $O(f_i,D_i)$, for any $f_i^1$ and ${f}_i^2$, there is:
\begin{equation*}\label{eq:theorem_3}
(f_i^1-{f}^2_i)^T(\nabla O(f_i^1,D_i)-\nabla O({f}^2_i,D_i)) \geq 0
\end{equation*}

Let $\langle\cdot,\cdot\rangle_F$ be frobenius inner product of two matrices, there is:
\begin{equation*}\label{eq:theorem_4}
\langle \hat{f}(t+1)-\hat{f}^*,\nabla \hat{O}(\hat{f}(t+1),D_{all})-\nabla \hat{O}(\hat{f}^*,D_{all})\rangle_F \geq 0
\end{equation*}

Substitute $\nabla\hat{O}(\hat{f}(t+1),D_{all})-\nabla \hat{O}(\hat{f}^*,D_{all})$ from \eqref{eq:theorem_1}:
\begin{equation}\label{eq:theorem_5}
\begin{gathered}
0 \leq \langle \hat{f}(t+1)-\hat{f}^*, -\sqrt{D-A} (Y(t+1)-Y^*)\rangle_F+\langle \hat{f}(t+1)-\hat{f}^*, -(W(t+1)-\theta I )(D-A)\hat{f}(t+1)\rangle_F\\+\langle \hat{f}(t+1)-\hat{f}^*,  - W(t+1)(D+A)(\hat{f}(t+1)-\hat{f}(t))\rangle_F 
\end{gathered}
\end{equation}

Consider the right hand side of \eqref{eq:theorem_5}. Since $D-A$ is symmetric and PSD, $\sqrt{D-A}$ is also a symmetric matrix and by \eqref{eq:theorem_2},
\begin{equation}\label{eq:theorem_6}
\begin{gathered}
\langle \hat{f}(t+1)-\hat{f}^*, -\sqrt{D-A} (Y(t+1)-Y^*)\rangle_F=\langle -\sqrt{D-A}(\hat{f}(t+1)-\hat{f}^*), (Y(t+1)-Y^*)\rangle_F\\ = -\langle \frac{1}{\theta}(Y(t+1)-Y(t)), Y(t+1)-Y^*\rangle_F
\end{gathered}
\end{equation}

Rearrange \eqref{eq:theorem_5} and use $(D-A)\hat{f}^* = \textbf{0}_{N\times d}$
\begin{equation}\label{eq:theorem_7}
\begin{gathered}
0 \geq\langle Z(t+1)-Z^*, J(t+1)(Z(t+1)-Z(t))\rangle_F + \langle \hat{f}(t+1)-\hat{f}^*, (W(t+1)-\theta I )(D-A)(\hat{f}(t+1)-\hat{f}^*)\rangle_F
\end{gathered}
\end{equation}

Suppose ${\eta}_i(t)\geq\theta$ for all $t,i$, {i.e.,} the diagonal matrix $W(t)-\theta I \succeq 0$ for all $t$. Since $D-A\succeq 0$, whose eigenvalues are all non-negative, the eigenvalues of $(W(t+1)-\theta I )(D-A)$ are thus also non-negative, {i.e.,} $(W(t+1)-\theta I )(D-A)\succeq 0$. Then for the second term of the RHS of \eqref{eq:theorem_7}, there is:
\begin{equation*}\label{eq:theorem_8}
\langle \hat{f}(t+1)-\hat{f}^*, (W(t+1)-\theta I )(D-A)(\hat{f}(t+1)-\hat{f}^*)\rangle_F \geq 0 
\end{equation*}

Therefore, 
\begin{equation}\label{eq:theorem_9}
\langle Z(t+1)-Z^*, J(t+1)(Z(t+1)-Z(t))\rangle_F \leq 0
\end{equation}

To simplify the notation, for a matrix $X$, let $||X||^2_{J} = \langle X, JX \rangle_F$, then \eqref{eq:theorem_9} can be represented as:  
\begin{equation*}\label{eq:theorem_10}
\begin{gathered}
\frac{1}{2}||Z(t+1)-Z^*||^2_{J(t+1)}+\frac{1}{2}||Z(t+1)-Z(t)||^2_{J(t+1)}-\frac{1}{2}||Z(t)-Z^*||^2_{J(t+1)}\leq 0
\end{gathered}
\end{equation*}

implies
\begin{equation}\label{eq:theorem_11}
\begin{gathered}
||Z(t+1)-Z(t)||^2_{J(t+1)}\leq -||Z(t+1)-Z^*||^2_{J(t+1)} +||Z(t)-Z^*||^2_{J(t)}+ ||Z(t)-Z^*||^2_{J(t+1)}-||Z(t)-Z^*||^2_{J(t)}
\end{gathered}
\end{equation}

Suppose ${\eta}_i(t+1)\geq{\eta}_i(t)$ for all $t$ and $i$, i.e., the diagonal matrix $W(t+1)-W(t)\succeq 0$ for all $t$. Since $D+A \succeq 0$, implies $(W(t+1)-W(t))(D+A) \succeq 0$. Let $U = \underset{i,t,k}{\text{sup}}|(f_i(t)-f_c^*)_k| \in \mathbb{R}$ be the finite upper bound of all nodes $i$, all iterations $t$ and all components $k$, then
\begin{equation}\label{eq:theorem_12}
\begin{gathered}
||Z(t)-Z^*||^2_{J(t+1)}-||Z(t)-Z^*||^2_{J(t)} = \text{Tr}((Z(t)-Z^*)^T(J(t+1)-J(t))(Z(t)-Z^*))
\\=  \text{Tr}((\hat{f}(t)-\hat{f}^*)^T(W(t+1)-W(t))(D+A)(\hat{f}(t)-\hat{f}^*)) \leq U^2(||\textbf{ones}(N,d)||^2_{W(t+1)(D+A)} - \textbf{ones}(N,d)||^2_{W(t)(D+A)})
\end{gathered}
\end{equation}

where $\textbf{ones}(N,d)$ is all one's matrix of size $N \times d$. By \eqref{eq:theorem_11}\eqref{eq:theorem_12}:
\begin{equation}\label{eq:theorem_13}
\begin{gathered}
||Z(t+1)-Z(t)||^2_{J(t+1)}\leq ||Z(t)-Z^*||^2_{J(t)}-||Z(t+1)-Z^*||^2_{J(t+1)}\\+U^2(||\textbf{ones}(N,d)||^2_{W(t+1)(D+A)} - ||\textbf{ones}(N,d)||^2_{W(t)(D+A)})
\end{gathered}
\end{equation}

Sum up \eqref{eq:theorem_13} over $t$ from $0$ to $+\infty$ leads to:
\begin{equation}\label{eq:theorem_14}
\begin{gathered}
\sum_{t=0}^{+\infty}||Z(t+1)-Z(t)||^2_{J(t+1)}\leq ||Z(0)-Z^*||^2_{J(0)}-||Z(+\infty)-Z^*||^2_{J(+\infty)}\\+U^2(||\textbf{ones}(N,d)||^2_{W(+\infty)(D+A)} - ||\textbf{ones}(N,d)||^2_{W(0)(D+A)})
\end{gathered}
\end{equation}

Since ${\eta}_i(t)<+\infty$, the RHS of \eqref{eq:theorem_14} is finite, implies that $\lim_{t\rightarrow+\infty}||Z(t+1)-Z(t)||^2_{J(t+1)} = 0$ must hold. 

By the definition of $Z(t)$, $J(t)$ and $||X||^2_{J} = \langle X, JX \rangle_F$, the following must hold  
\begin{equation}\label{eq:theorem_15}
\lim_{t\rightarrow+\infty}||\hat{f}(t+1)-\hat{f}(t)||^2_{W(t+1)(D+A)} = 0
\end{equation}
\begin{equation}\label{eq:theorem_16}
\lim_{t\rightarrow+\infty}||Y(t+1)-Y(t)||^2_{F} = 0
\end{equation}

\eqref{eq:theorem_16} shows that $Y(t)$ converges to a stationary point ${Y}^{s}$, along with \eqref{eq:converge_8} imply $\lim_{t\rightarrow+\infty} \sqrt{D-A}\hat{f}(t+1) = 0$. Since $\text{Null}( \sqrt{D-A}) = c\textbf{1}$, $\hat{f}(t+1)$ must lie in the subspace spanned by $\textbf{1}$ as $t \rightarrow \infty$. To satisfy \eqref{eq:theorem_15}, either of the following two statements must hold:
\begin{itemize}
\item $\lim_{t\rightarrow+\infty}(\hat{f}(t+1)-\hat{f}(t)) = \textbf{0}_{N\times d}$
\item $\lim_{t\rightarrow+\infty}W(t+1)(D+A)\textbf{1}= \lim_{t\rightarrow+\infty}W(t+1)A\textbf{1}+\lim_{t\rightarrow+\infty}\sum_{i=1}^{N}{\eta}_i(t+1)V_i=\textbf{0}_{N\times 1}$
\end{itemize}

Since ${\eta}_i(t) \geq \theta > 0$ for all $t$, implies $\lim_{t\rightarrow+\infty}\sum_{i=1}^{N}{\eta}_i(t+1)V_i>0$. The second statement can never be true because all elements of $A$ and $W(t+1)$ are non-negative. Hence, $\hat{f}(t)$ should also converge to a stationary point $\hat{f}^{s}$.

Now show that the stationary point $({Y}^s,\hat{f}^s)$ is $(Y^*,\hat{f}^*)$.

Take limit of both sides of \eqref{eq:converge_7} \eqref{eq:converge_8}, substitute $\hat{f}^s,{Y}^s$ yields 

\begin{equation}\label{eq:theorem_17}
\begin{gathered}
\nabla \hat{O}(\hat{f}^s,D_{all}) + \sqrt{D-A}{Y}^s+(W(t+1)-\theta I )(D-A)\hat{f}^s =\textbf{0}_{N\times d}
\end{gathered}
\end{equation}
\begin{equation}\label{eq:theorem_18}
\sqrt{D-A}\hat{f}^s=\textbf{0}_{N\times d}
\end{equation}

By \eqref{eq:theorem_18}, \eqref{eq:theorem_17} turns into:
\begin{equation}\label{eq:theorem_19}
\begin{gathered}
\nabla \hat{O}(\hat{f}^s,D_{all}) + \sqrt{D-A}{Y}^s =\textbf{0}_{N\times d}
\end{gathered}
\end{equation}

Compare \eqref{eq:theorem_18}\eqref{eq:theorem_19}
with \eqref{eq:converge_9}\eqref{eq:converge_10} in Lemma \ref{Lemma:1} and observe that $({Y}^s,\hat{f}^s)$ satisfies the optimality condition \eqref{eq:converge_9}\eqref{eq:converge_10} and is thus the optimal point. Therefore, $f(t)$ converges to $\hat{f}^*$ and $Y(t)$ converges to $Y^*$.

\section{Proof of Theorem \ref{Theorem:2}}

According to the Assumption 3 that $O(f_i,D_i)$ is strongly convex and has Lipschitz continues gradients for all $i \in \mathscr{N}$, define diagonal matrices $D_m = \textbf{diag}([m_1;m_2;\cdots;m_N]) \in \mathbb{R}^{N \times N}$ and $D_M = \textbf{diag}([M_1^2;M_2^2;\cdots;M_N^2])\in \mathbb{R}^{N \times N}$, \eqref{eq:converge_1} yield: 

\begin{equation}\label{eq:lemma2_1}
\begin{gathered}
\langle \hat{f}^1 - \hat{f}^2,\nabla \hat{O}(\hat{f}^1,D_{all}) -\nabla \hat{O}(\hat{f}^2,D_{all}) \rangle_F\geq \langle \hat{f}^1- \hat{f}^2,D_m(\hat{f}^1 - \hat{f}^2) \rangle_F
\end{gathered}
\end{equation}
 \begin{equation}\label{eq:lemma2_2}
 \begin{gathered}
||\nabla \hat{O}(\hat{f}^1,D_{all}) -\nabla \hat{O}(\hat{f}^2,D_{all})||^2_F \leq \langle \hat{f}^1 - \hat{f}^2,D_M(\hat{f}^1 - \hat{f}^2) \rangle_F 
 \end{gathered}
\end{equation}

Since for any $\mu > 1$ and any matrices $C_1$, $C_2$ with the same dimensions, there is:
\begin{equation*}\label{eq:lemma2_7}
||C_1+C_2||^2_F \leq \mu||C_1||^2_F+ \frac{\mu}{\mu - 1}||C_2||^2_F
\end{equation*}

From \eqref{eq:theorem_1}, there is:
\begin{equation}\label{eq:lemma2_8}
\begin{gathered}
||\sqrt{D-A} (Y(t+1)-Y^*)||^2_F \leq  \mu ||\nabla \hat{O}(\hat{f}(t+1),D_{all})- \nabla\hat{O}(\hat{f}^*,D_{all})+ W(t+1)(D+A)(\hat{f}(t+1)-\hat{f}(t))||^2_F\\ + \frac{\mu}{\mu -1}||(W(t+1)-\theta I )(D-A)\hat{f}(t+1)||^2_F \leq \frac{\mu^2}{\mu-1}||\nabla \hat{O}(\hat{f}(t+1),D_{all})- \nabla\hat{O}(\hat{f}^*,D_{all})||_F^2 \\+\mu^2||W(t+1)(D+A)(\hat{f}(t+1)-\hat{f}(t))||_F^2+ \frac{\mu}{\mu -1}||(W(t+1)-\theta I )(D-A)\hat{f}(t+1)||^2_F
\end{gathered}
\end{equation}

Let $\sigma_{\text{min}}(\cdot)$, $\sigma_{\text{max}}(\cdot)$ denote the smallest nonzero singular value and the largest singular value of a matrix respectively.

For any matrices $C_1$, $C_2$, let $C_1 = U\Sigma V^T$ be SVD of $C_1$, there is:
\begin{equation*}\label{eq:lemma2_9}
|| C_1C_2||_F^2 \leq \sigma_{\text{max}}(C_1) || C_2||_{C_1^T}^2
\end{equation*}
\begin{equation*}\label{eq:lemma2_10}
\begin{gathered}
\sigma_{\text{min}}(C_1)^2 || C_2||_F^2 \leq || C_1C_2||_F^2 \leq \sigma_{\text{max}}(C_1)^2 || C_2||_F^2
\end{gathered}
\end{equation*}

Denote $$\bar{\sigma}_{\text{max}}(t+1) = \sigma_{\text{max}}({(W(t+1)-\theta I )(D-A)})$$  $$\bar{\sigma}_{\text{min}}(t+1) = \sigma_{\text{min}}({(W(t+1)-\theta I )(D-A)})$$ 
$$\tilde{\sigma}_{\text{max}}(t+1)=\sigma_{\text{max}}({W(t+1)(D+A)})$$

Using \eqref{eq:lemma2_2} and $(D-A)\hat{f}^* = 0$, \eqref{eq:lemma2_8} is turned into:
\begin{equation*}\label{eq:lemma2_11}
\begin{gathered}
\frac{1}{\theta}||Y(t+1)-Y^*||^2_F  \leq\frac{\mu^2}{\theta\sigma_{\text{min}}(D-A)(\mu-1)}||\hat{f}(t+1)-\hat{f}^*||_{D_M}^2\\+  \frac{\mu^2\tilde{\sigma}_{\text{max}}(t+1)}{\theta\sigma_{\text{min}}(D-A)}||\hat{f}(t+1)-\hat{f}(t)||_{W(t+1)(D+A)}^2+ \frac{\mu\bar{\sigma}_{\text{max}}(t+1)^2}{\theta\sigma_{\text{min}}(D-A)(\mu -1)}||(\hat{f}(t+1)-\hat{f}^*)||^2_{F}
\end{gathered}
\end{equation*}

Adding $||\hat{f}(t+1)-\hat{f}^*||^2_{W(t+1)(D+A)}$ at both sides leads to:
\begin{equation}\label{eq:lemma3_2}
\begin{gathered}
||Z(t+1)-Z^*||^2_{J(t+1)}  \leq \frac{\mu^2\tilde{\sigma}_{\text{max}}(t+1)}{\theta\sigma_{\text{min}}(D-A)}||\hat{f}(t+1)-\hat{f}(t)||_{W(t+1)(D+A)}^2\\+||\hat{f}(t+1)-\hat{f}^*||_{\frac{\mu^2D_M+\mu\bar{\sigma}_{\text{max}}(t+1)^2\textbf{I}_N}{\theta\sigma_{\text{min}}(D-A)(\mu-1)}+W(t+1)(D+A)}^2
\end{gathered}
\end{equation}

Since 
\begin{equation}\label{eq:delta1}
\frac{\delta(t+1)\mu^2\tilde{\sigma}_{\text{max}}(t+1)}{\theta\sigma_{\text{min}}(D-A)}\leq 1
\end{equation}

and 
\begin{equation}\label{eq:delta2}
\begin{gathered}
\delta(t+1)(\frac{\mu\bar{\sigma}_{\text{max}}(t+1)^2\textbf{I}_N+\mu^2D_M}{\theta\sigma_{\text{min}}(D-A)(\mu-1)}+W(t+1)(D+A)) \preceq  2(W(t+1)-\theta I )(D-A) + 2D_m
\end{gathered}
\end{equation}

It implies from \eqref{eq:lemma3_2} that:
\begin{equation}\label{eq:lemma}
\begin{gathered}
\delta(t+1)||Z(t+1)-Z^*||^2_{J(t+1)}  \leq||\hat{f}(t+1)-\hat{f}(t)||_{W(t+1)(D+A)}^2+||\hat{f}(t+1)-\hat{f}^*||^2_{2(W(t+1)-\theta I )(D-A) + 2D_m}\\
\leq ||Z(t+1)-Z(t)||^2_{J(t+1)}+||\hat{f}(t+1)-\hat{f}^*||^2_{2(W(t+1)-\theta I )(D-A) + 2D_m}
\end{gathered}
\end{equation}

Substituting $\hat{f}^1$ with $\hat{f}(t+1)$ and $\hat{f}^2$ with $\hat{f}^*$ and the gradient difference from \eqref{eq:theorem_1} in \eqref{eq:lemma2_1} leads to:
\begin{equation*}\label{eq:lemma2_3}
\begin{gathered}
\langle \hat{f}(t+1) - \hat{f}^*,\sqrt{D-A} (Y(t+1)-Y^*) \rangle_F+ \langle \hat{f}(t+1) - \hat{f}^*,W(t+1)(D+A)(\hat{f}(t+1)-\hat{f}(t)) \rangle_F \\ +\langle \hat{f}(t+1) - \hat{f}^*,(W(t+1)-\theta I )(D-A)\hat{f}(t+1)  \rangle_F  \leq -\langle \hat{f}(t+1) - \hat{f}^*,D_m(\hat{f}(t+1)  - \hat{f}^*) \rangle_F
\end{gathered}
\end{equation*}

Similar to the proof of Theorem \ref{Theorem:1},
using the definition of $Z(t+1)$, $Z^*$, $J(t+1)$ and $(D-A)\hat{f}^* = 0$, there is:
\begin{equation}\label{eq:lemma2_5}
\begin{gathered}
||Z(t+1)-Z^*||^2_{J(t+1)}\leq -||Z(t+1)-Z(t)||^2_{J(t+1)}+||Z(t)-Z^*||^2_{J(t+1)} - || \hat{f}(t+1)-\hat{f}^*||^2_{2D_m+2(W(t+1)-\theta I )(D-A)}
\end{gathered}
\end{equation}

Sum up \eqref{eq:lemma} and \eqref{eq:lemma2_5} gives:
\begin{equation*}
(1+\delta(t+1))||Z(t+1)-Z^*||^2_{J(t+1)} \leq ||Z(t)-Z^*||^2_{J(t+1)}
\end{equation*}

Let $m_o = \min_{i \in \mathscr{N}}\{m_i\}$, $M_O = \max_{i \in \mathscr{N}}\{M_i\}$. One $\delta(t+1)$ that satisfies \eqref{eq:delta1} and \eqref{eq:delta2} could be:
\begin{equation*}
\begin{gathered}
\min \{\frac{\theta\sigma_{\text{min}}(D-A)}{\mu^2\tilde{\sigma}_{\text{max}}(t+1)}, \frac{2m_o + 2\bar{\sigma}_{\text{min}}(t+1)}{\frac{\mu^2M_O^2+\mu\bar{\sigma}_{\text{max}}(t+1)^2}{\theta\sigma_{\text{min}}(D-A)(\mu-1)} + \tilde{\sigma}_{\text{max}}(t+1)}\}
\end{gathered}
\end{equation*}

\section{Proof of Theorem \ref{thmP}}
In the following proof, use the uppercase letters and lowercase letters to denote random variables and the corresponding realizations.

Since the modified ADMM is randomized, denote $F_i(t)$ as the random variable of the result that node $i$ broadcasts in $t$-th iteration, of which the realization is $f_i(t)$. Define $F(t) = \{F_i(t)\}_{i=1}^N$ whose realization is $\{f_i(t)\}_{i=1}^N$.

Let $\mathscr{F}_{F(0:t)}(\cdot)$ be the joint probability distribution of $F(0:t)=\{F(r)\}_{r=0}^t$, and $\mathscr{F}_{F(t)}(\cdot)$ be the distribution of $F(t)$, by chain rule:
\begin{equation*}\label{P_analysis_1}
\begin{gathered}
\mathscr{F}_{F(0:T)}(\{f(r)\}_{r=0}^T) = \mathscr{F}_{F(0:T-1)}(\{f(r)\}_{r=0}^{T-1}) \cdot \mathscr{F}_{F(T)}(f(T)|\{f(r)\}_{r=0}^{T-1})
=\cdots \\= \mathscr{F}_{F(0)}(f(0))\cdot \prod^T_{t=1}\mathscr{F}_{F(t)}(f(t)|\{f(r)\}_{r=0}^{t-1})
\end{gathered}
\end{equation*}

For two neighboring datasets $D_{all}$ and $\hat{D}_{all}$ of the network, the ratio of joint probabilities is given by:
\begin{equation}\label{P_analysis_2}
\begin{gathered}
\frac{\mathscr{F}_{F(0:T)}(\{f(r)\}_{r=0}^T|D_{all})}{\mathscr{F}_{F(0:T)}(\{f(r)\}_{r=0}^T|\hat{D}_{all})} = \frac{\mathscr{F}_{F(0)}(f(0)|D_{all})}{\mathscr{F}_{F(0)}(f(0)|\hat{D}_{all})}  \cdot \prod^T_{t=1}\frac{\mathscr{F}_{F(t)}(f(t)|\{f(r)\}_{r=0}^{t-1},D_{all})}{\mathscr{F}_{F(t)}(f(t)|\{f(r)\}_{r=0}^{t-1},\hat{D}_{all})}
\end{gathered}
\end{equation}

Since $f_i(0)$ is randomly selected for all $i$, which is independent of dataset, there is $\mathscr{F}_{F(0)}(f(0)|D_{all}) = \mathscr{F}_{F(0)}(f(0)|\hat{D}_{all})$.

First only consider $t$-th iteration, since the primal variable is updated according to \eqref{eq:P_modify_2}, by KKT optimality condition, $\nabla_{f_i}{L}_i^{priv} (t)|_{f_i = f_i(t)} = 0$, implies:
\begin{equation}\label{P_analysis_3}
\begin{gathered}
\epsilon_i(t) = -\frac{1}{2\eta_i(t)V_i}\frac{C}{B_i}\sum_{n=1}^{B_i}y^n_i\mathscr{L}'(y_i^n f_i(t)^T x_i^n)x_i^n
-\frac{1}{2\eta_i(t)V_i}(\frac{\rho}{N}\nabla R(f_i(t)) + 2 \lambda_i(t-1))
\\-\frac{1}{2V_i}\sum_{j \in \mathscr{V}_i}(2f_i(t) - f_i(t-1) - f_j(t-1))
\end{gathered}
\end{equation}

Given $\{f_i(r)\}_{r=0}^{t-1}$, $F_i(t)$ and $E_i(t)$ will be bijective:
\begin{itemize}
\item For any $F_i(t)$ with the realization $f_i(t)$, $\exists$ an unique $E_i(t)=\epsilon_i(t)$ having the form of \eqref{P_analysis_3} such that the KKT condition holds.
\item Since the Lagrangian $L_i^{priv}(t)$ is strictly convex (by Assumption 4,5), its minimizer is unique, implies that for any $E_i(t)$ with the realization $\epsilon_i(t)$, $\exists$ an unique $F_i(t)=f_i(t)$ such that the KKT condition holds.
\end{itemize} 

Since each node $i$ generates $\epsilon_i(t)$ independently, $f_i(t)$ is also independent from each other. Let $\mathscr{F}_{F_i(t)}(\cdot)$ be the distribution of $F_i(t)$, there is:
\begin{equation}\label{P_analysis_4}
\begin{gathered}
\frac{\mathscr{F}_{F(t)}(f(t)|\{f(r)\}_{r=0}^{t-1},D_{all})}{\mathscr{F}_{F(t)}(f(t)|\{f(r)\}_{r=0}^{t-1},\hat{D}_{all})}
= \prod^{N}_{v=1}\frac{\mathscr{F}_{F_v(t)}(f_v(t)|\{f_v(r)\}_{r=0}^{t-1},D_v)}{\mathscr{F}_{F_v(t)}(f_v(t)|\{f_v(r)\}_{r=0}^{t-1},\hat{D}_v)}
= \frac{\mathscr{F}_{F_i(t)}(f_i(t)|\{f_i(r)\}_{r=0}^{t-1},D_i)}{\mathscr{F}_{F_i(t)}(f_i(t)|\{f_i(r)\}_{r=0}^{t-1},\hat{D}_i)}
\end{gathered}
\end{equation}

Since two neighboring datasets $D_{all}$ and $\hat{D}_{all}$ only have at most one data point that is different, the second equality holds is because of the fact that this different data point could only be possessed by one node, say node $i$. Then there is $D_j = \hat{D}_j$ for $j \neq i$.

Given $\{f_i(r)\}_{r=0}^{t-1}$, let $g_{t}(\cdot,D_i): \mathbb{R}^d \rightarrow \mathbb{R}^d $ denote the one-to-one mapping from $E_i(t)$ to $F_i(t)$ using dataset $D_i$. Let $\mathscr{F}_{E_i(t)}(\cdot)$ be the probability density of $E_i(t)$, by Jacobian transformation, there is\footnote{We believe that there is a critical mistake in \cite{zhang2017} and the original paper \cite{chaudhuri2011} where the objective perturbation method was proposed. A wrong mapping is used in both work:\begin{equation*}
\begin{gathered}
\mathscr{F}_{F_i(t)}(f_i(t)|D_i) = \mathscr{F}_{E_i(t)}(g^{-1}_{t}(f_i(t),D_i)) \cdot|\det(\textbf{J}(g^{-1}_{t}(f_i(t),D_i)))|^{-1}
\end{gathered}
\end{equation*}}:
\begin{equation}\label{P_analysis_5}
\begin{gathered}
\mathscr{F}_{F_i(t)}(f_i(t)|D_i) = \mathscr{F}_{E_i(t)}(g^{-1}_{t}(f_i(t),D_i)) \cdot|\det(\textbf{J}(g^{-1}_{t}(f_i(t),D_i)))|
\end{gathered}
\end{equation}
where $g^{-1}_{t}(f_i(t),D_i)$ is the mapping from $F_i(t)$ to $E_i(t)$ using data $D_i$ as shown in \eqref{P_analysis_3} and $\textbf{J}(g^{-1}_{t}(f_i(t),D_i))$ is the Jacobian matrix of it.

Without loss of generality, let $D_i$ and $\hat{D}_i$ be only different in the first data point, say $(x_i^1,y_i^1)$ and $(\hat{x}_i^1,\hat{y}_i^1)$ respectively. Then by \eqref{P_analysis_4}\eqref{P_analysis_5}, \eqref{P_analysis_2} yields:
\begin{equation}\label{P_analysis_6}
\begin{gathered}
\frac{\mathscr{F}_{F(0:T)}(\{f(r)\}_{r=0}^T|D_{all})}{\mathscr{F}_{F(0:T)}(\{f(r)\}_{r=0}^T|\hat{D}_{all})}= \prod^{T}_{t=1}\frac{\mathscr{F}_{E_i(t)}(g^{-1}_{t}(f_i(t),D_i))}{\mathscr{F}_{E_i(t)}(g^{-1}_{t}(f_i(t),\hat{D}_i))}
\cdot \prod^{T}_{t=1} \frac{|\det(\textbf{J}(g^{-1}_{t}(f_i(t),D_i)))|}{|\det(\textbf{J}(g^{-1}_{t}(f_i(t),\hat{D}_i)))|}
\end{gathered}
\end{equation}

Consider the first part, $E_i(t) \sim \exp\{-\alpha_i(t)||\epsilon||\}$, let $\hat{\epsilon}_i(t) = g^{-1}_{t}(f_i(t),\hat{D}_i)$ and ${\epsilon}_i(t) = g^{-1}_{t}(f_i(t),D_i)$
\begin{equation}\label{P_analysis_7}
\begin{gathered}
\prod^{T}_{t=1}\frac{\mathscr{F}_{E_i(t)}(g^{-1}_{t}(f_i(t),D_i))}{\mathscr{F}_{E_i(t)}(g^{-1}_{t}(f_i(t),\hat{D}_i))}
= \prod^{T}_{t=1} \exp(\alpha_i(t)(||\hat{\epsilon}_i(t)|| - ||\epsilon_i(t)||))
\leq \exp(\sum^{T}_{t=1}\alpha_i(t)||\hat{\epsilon}_i(t) - \epsilon_i(t)||)
\end{gathered}
\end{equation}

By \eqref{P_analysis_3}, Assumptions 4 and the facts that $||x_i^n||_2 \leq 1$ (pre-normalization), $y_i^n \in \{+1,-1\}$.
\begin{equation*}\label{P_analysis_8}
\begin{gathered}
||\hat{\epsilon}_i(t) - \epsilon_i(t)|| = \frac{1}{2\eta_i(t)V_i}\frac{C}{B_i} \cdot ||y^1_i\mathscr{L}'(y_i^1 f_i(t)^T x_i^1)x_i^1 - \hat{y}^1_i\mathscr{L}'(\hat{y}_i^1 f_i(t)^T \hat{x}_i^1)\hat{x}_i^1|| \leq \frac{C}{\eta_i(t)V_iB_i}
\end{gathered}
\end{equation*}

\eqref{P_analysis_7} can be bounded:
\begin{equation}\label{P_analysis_9}
\begin{gathered}
\prod^{T}_{t=1}\frac{\mathscr{F}_{E_i(t)}(g^{-1}_{t}(f_i(t),D_i))}{\mathscr{F}_{E_i(t)}(g^{-1}_{t}(f_i(t),\hat{D}_i))}
 \leq \exp(\sum^{T}_{t=1}\frac{C\alpha_i(t)}{\eta_i(t)V_iB_i})
\end{gathered}
\end{equation}

Consider the second part, the Jacobian matrix $\textbf{J}(g^{-1}_{t}(f_i(t),D_i))$ is:
\begin{equation*}\label{P_analysis_10}
\begin{gathered}
\textbf{J}(g^{-1}_{t}(f_i(t),D_i)) = -\frac{1}{2\eta_i(t)V_i}\frac{C}{B_i}\sum_{n=1}^{B_i}\mathscr{L}''(y_i^n f_i(t)^T x_i^n)x_i^n(x_i^n)^T
-\frac{1}{2\eta_i(t)V_i}\frac{\rho}{N}\nabla^2 R(f_i(t)) - \textbf{I}_d
\end{gathered}
\end{equation*}

Let $G(t) = \frac{C}{2\eta_i(t)V_iB_i}(\mathscr{L}''(\hat{y}_i^1 f_i(t)^T \hat{x}_i^1)\hat{x}_i^1(\hat{x}_i^1)^T - \mathscr{L}''(y_i^1 f_i(t)^T x_i^1)x_i^1(x_i^1)^T)$ and $H(t) = -\textbf{J}(g^{-1}_{t}(f_i(t),D_i))$, there is:
\begin{equation*}\label{P_analysis_10}
\begin{gathered}
 \frac{|\det(\textbf{J}(g^{-1}_{t}(f_i(t),D_i)))|}{|\det(\textbf{J}(g^{-1}_{t}(f_i(t),\hat{D}_i)))|}
 = \frac{|\det(H(t))|}{|\det(H(t)+G(t))|}
 = \frac{1}{|\det(I + H(t)^{-1}G(t))|}
 = \frac{1}{|\prod_{j=1}^r(1+\lambda_j(H(t)^{-1}G(t)))|}
\end{gathered}
\end{equation*}

where $\lambda_j(H(t)^{-1}G(t))$ denotes the $j$-th largest eigenvalue of $H(t)^{-1}G(t)$. Since $G(t)$ has rank at most 2, implies $H(t)^{-1}G(t)$ also has rank at most 2. 

Because $\theta$ is determined such that $2c_1 < \frac{B_i}{C}(\frac{\rho}{N}+2\theta V_i)$, and $\theta \leq \eta_i(t)$ holds for all node $i$ and iteration $t$, which implies:
\begin{equation}\label{eq:condition}
\frac{c_1}{\frac{B_i}{C}(\frac{\rho}{N}+2\eta_i(t)V_i)} < \frac{1}{2}
\end{equation}

By Assumptions 4 and 5, the eigenvalue of $H(t)$ and $G(t)$ satisfy:
\begin{equation*}\label{P_analysis_12}
\lambda_j(H(t)) \geq \frac{\rho}{2\eta_i(t)V_iN} + 1 > 0 
\end{equation*}
\begin{equation*}\label{P_analysis_13}
-\frac{Cc_1}{2\eta_i(t)V_iB_i} \leq \lambda_j(G(t)) \leq \frac{Cc_1}{2\eta_i(t)V_iB_i}
\end{equation*}

Implies:
\begin{equation*}\label{P_analysis_14}
\begin{gathered}
-\frac{c_1}{\frac{B_i}{C}(\frac{\rho}{N}+2\eta_i(t)V_i)}\leq \lambda_{j}(H(t)^{-1}G(t))
\leq \frac{c_1}{\frac{B_i}{C}(\frac{\rho}{N}+2\eta_i(t)V_i)}
\end{gathered}
\end{equation*}

By \eqref{eq:condition}:
\begin{equation*}
\begin{gathered}
-\frac{1}{2}\leq \lambda_{j}(H(t)^{-1}G(t))
\leq \frac{1}{2}
\end{gathered}
\end{equation*}

Since $\lambda_{\min}(H(t)^{-1}G(t)) > -1$, there is:
\begin{equation*}\label{P_analysis_11}
\begin{gathered}
\frac{1}{|1+\lambda_{\max}(H(t)^{-1}G(t))|^2}  \leq \frac{1}{|\text{det}(I+H(t)^{-1}G(t))|}  \leq \frac{1}{|1+\lambda_{\min}(H(t)^{-1}G(t))|^2}
\end{gathered}
\end{equation*}

Therefore, 
\begin{equation}\label{P_analysis_15}
\begin{gathered}
 \prod^{T}_{t=1}\frac{|\det(\textbf{J}(g^{-1}_{t}(f_i(t),D_i)))|}{|\det(\textbf{J}(g^{-1}_{t}(f_i(t),\hat{D}_i)))|}
 \leq \prod^{T}_{t=1}\frac{1}{|1-\frac{c_1}{\frac{B_i}{C}(\frac{\rho}{N}+2\eta_i(t)V_i)}|^2}
 = \exp(-\sum_{t=1}^{T}2\ln(1-\frac{c_1}{\frac{B_i}{C}(\frac{\rho}{N}+2\eta_i(t)V_i)}))
\end{gathered}
\end{equation}

Since for any real number $x \in [0,0.5]$, $-\ln(1-x)<1.4x$. By condition \eqref{eq:condition}, \eqref{P_analysis_15} can be bounded with a simper expression:
\begin{equation}\label{P_analysis_16}
\begin{gathered}
 \prod^{T}_{t=1}\frac{|\det(\textbf{J}(g^{-1}_{t}(f_i(t),D_i)))|}{|\det(\textbf{J}(g^{-1}_{t}(f_i(t),\hat{D}_i)))|}
 \leq \exp(\sum_{t=1}^{T}\frac{2.8c_1}{\frac{B_i}{C}(\frac{\rho}{N}+2\eta_i(t)V_i)}) \leq \exp(\sum_{t=1}^{T}\frac{1.4Cc_1}{\eta_i(t)V_iB_i})
\end{gathered}
\end{equation}

Combine \eqref{P_analysis_9}\eqref{P_analysis_16}, \eqref{P_analysis_6} can be bounded:
\begin{equation*}
\begin{gathered}
\frac{\mathscr{F}_{F(0:T)}(\{f(r)\}_{r=0}^T|D_{all})}{\mathscr{F}_{F(0:T)}(\{f(r)\}_{r=0}^T|\hat{D}_{all})}  \leq\exp(\sum_{t=1}^{T}(\frac{1.4Cc_1}{\eta_i(t)V_iB_i}+\frac{C\alpha_i(t)}{\eta_i(t)V_iB_i}))
= \exp(\sum_{t=1}^{T}\frac{C}{\eta_i(t)V_iB_i}(1.4c_1+\alpha_i(t)))
\end{gathered}
\end{equation*}

Therefore, the total privacy loss during $T$ iterations can be bounded by any $\beta$:
\begin{equation*}
 \beta \geq \underset{i \in \mathscr{N}}{\max}\{\sum_{t=1}^{T}\frac{C}{\eta_i(t)V_iB_i}(1.4c_1+\alpha_i(t))\}
\end{equation*}

\section{Inference of Attackers when $\eta_i(t)$ is Non-private}
By KKT optimality condition in each iteration, we have:
\begin{eqnarray}\label{eq:dis_2}
\epsilon_i(t)+\frac{1}{2\eta_i(t) V_i}\frac{C}{B_i}y^1_i\mathscr{L}'(y_i^1 f_i(t)^T x_i^1)x_i^1 \nonumber = -\frac{1}{2\eta_i(t) V_i}\frac{C}{B_i}\sum_{n=2}^{B_i}y^n_i\mathscr{L}'(y_i^n f_i(t)^T x_i^n)x_i^n\\
-\frac{1}{2\eta_i(t) V_i}(\frac{\rho}{N}\nabla R(f_i(t)) + 2 \lambda_i(t-1))\nonumber
-\frac{1}{2V_i}\sum_{j \in \mathscr{V}_i}(2f_i(t) - f_i(t-1) - f_j(t-1))~.\nonumber
\end{eqnarray}
In this case the attacker can compute the RHS of \eqref{eq:dis_2} completely.  Furthermore, since $E_i(t)$ is zero-mean, over a large number of iterations we will have $\frac{1}{T}\sum_{t=1}^{T} \epsilon_i(t) \approx 0$ with high probability, which then allows the attacker to determine the features of the unknown individual up to a scaling factor, i.e., it can determine the second term on the LHS as a scalar multiplied with $x_i^1$.

\end{document}